\documentclass{article}

\usepackage[nonatbib,final]{neurips_2024}

\usepackage[utf8]{inputenc} %
\usepackage[T1]{fontenc}    %
\usepackage{hyperref}       %
\usepackage{url}            %
\usepackage{booktabs}       %
\usepackage{amsfonts}       %
\usepackage{nicefrac}       %
\usepackage{microtype}      %
\usepackage{xcolor}         %

\usepackage{amsmath,amsfonts}
\usepackage{algorithmic}
\usepackage{algorithm}
\usepackage{array}

\usepackage{textcomp}
\usepackage{stfloats}
\usepackage{url}
\usepackage{verbatim}
\usepackage{graphicx}

\usepackage{multirow}
\usepackage{booktabs}
\usepackage{arydshln}

\usepackage{caption}
\usepackage{subcaption}

\usepackage{amssymb}
\usepackage{amsbsy}
\usepackage{pifont}

\usepackage{wrapfig}
\usepackage{tcolorbox}

\usepackage{hyperref}

\definecolor{citecolor}{HTML}{0071bc}
\definecolor{paleplum}{rgb}{0.8, 0.6, 0.8}
\hypersetup{
  colorlinks,
  citecolor=citecolor,
  linkcolor=red
}

\newcommand{\cmark}{\ding{51}}
\newcommand{\xmark}{\ding{55}}

\DeclareMathOperator*{\argmin}{arg\,min}

\tcbset{colback=yellow!10!white, colframe=red!50!black, width=\textwidth, arc=0mm, boxrule=1mm}

\title{Point-PRC: A Prompt Learning Based Regulation Framework for Generalizable Point Cloud Analysis}

\author{%
  Hongyu Sun\textsuperscript{1,2} 
  \quad Qiuhong Ke\textsuperscript{2} 
  \quad Yongcai Wang\textsuperscript{1}\thanks{Corresponding author. $\tt \{sunhongyu, ycw\}@ruc.edu.cn, \{qiuhong.ke, jianfei.cai\}@monash.edu$}\\[0.1cm]
  \quad \textbf{Wang Chen}\textsuperscript{1} 
  \quad \textbf{Kang Yang}\textsuperscript{1} 
  \quad \textbf{Deying Li}\textsuperscript{1} 
  \quad \textbf{Jianfei Cai}\textsuperscript{2} \\[0.5cm]
  \textsuperscript{1}Department of Computer Science, Renmin University of China, China \\[0.1cm]
  \textsuperscript{2}Department of Data Science \& AI, Monash University, Australia
}

\begin{document}

\maketitle

\begin{abstract}
This paper investigates the 3D domain generalization (3DDG) ability of large 3D models based on prevalent prompt learning. 
Recent works demonstrate the performances of 3D point cloud recognition can be boosted remarkably by parameter-efficient 
prompt tuning. However, we observe that the improvement on downstream tasks comes at the expense of a severe drop 
in 3D domain generalization. To resolve this challenge, we present a comprehensive regulation framework that 
allows the learnable prompts to actively interact with the well-learned general knowledge in large 3D models to 
maintain good generalization. Specifically, the proposed framework imposes multiple explicit constraints on 
the prompt learning trajectory by maximizing the mutual agreement between task-specific predictions and task-agnostic knowledge.  
We design the regulation framework as a plug-and-play module to embed into existing representative large 3D models. 
Surprisingly, our method not only realizes consistently increasing generalization ability but also enhances task-specific 
3D recognition performances across various 3DDG benchmarks by a clear margin. Considering the lack of study and evaluation 
on 3DDG, we also create three new benchmarks, namely base-to-new, cross-dataset and few-shot generalization benchmarks, 
to enrich the field and inspire future research. Code and benchmarks are available at \url{https://github.com/auniquesun/Point-PRC}.
\end{abstract}

\section{Introduction}
\label{sec:intro}
3D point cloud data is widely adopted in many industrial and civil areas, such as autonomous driving~\cite{qi19votenet}, 
robotics~\cite{levinson10robust,behley18efficient}, geospatial mapping~\cite{mccurley01geospatial} and entertainment games~\cite{placitelli11low}. 
Recognizing 3D objects from point cloud data is a basic need of these applications. 
Relevant research topics have been explored for a long time and their development can be summarized in three stages. 
In the early phase, PointNet series~\cite{qi17pointnet,qi17pointnet2} sparked a wave of directly operating raw point cloud data using 
deep learning techniques. 
Later methods improved upon PointNet and PointNet++ in terms of local information 
aggregation~\cite{li18pointcnn,wu19pointconv,thomas19kpconv,wang19dgcnn,xiang21curvenet,ma22pointmlp}, 
optimization techniques~\cite{qian22pointnext}, geometry prior~\cite{ran22repsurf}, 
model architecture~\cite{guo2020pct,zhao21pt,wu22ptv2,park23spotr,duan23condaformer}, \textit{etc}. 
Although remarkable progress has been made, these works tend to design specific architectures targeting downstream 
benchmarks while paying little attention to the model generalization, resulting in disappointed performances when 
deploying in the wild, especially in the case of unseen domains and corrupted data. 
On the other hand, training point cloud recognition models on each benchmark is not always feasible due to 
the narrow set of 3D visual concepts and expensive labeled data. 

The above factors call for the investigation of the domain generalization routes for the deep point cloud models
so that they can learn robust and transferable representations. 
Related studies have been extensively conducted in image 
recognition~\cite{li17dg,li18dg_metalearning,li18dg_adversarial,li19episodic,li20dg_medical,zhou21dg_mixstyle,ding22dg,zhou23dg} 
while to our best knowledge, there are only a few methods to discuss the domain adaptation and domain generalization in 3D. 
Several years ago, 
PointDAN~\cite{qin19pointdan} first investigated domain adaptation for point cloud classification models by aligning 
multi-scale features of 3D objects across the source and target domains. 
MetaSets~\cite{huang21metasets} proposed to meta-learn on a group of transformed point sets to obtain generalizable 
representations to handle the sim-to-real geometry shifts. 
PDG~\cite{wei22pdg} decomposed 3D objects into shared part space to reduce domain gap and developed a part-level domain 
generalization model for 3D point cloud classification.

However, the above methods are all built on small models (e.g., PointNet with 1.2M parameters) and small datasets 
(e.g., ModelNet with 9,843 training samples) and the overall 
\begin{wrapfigure}{r}{0.55\linewidth}
    \centering
    \includegraphics[width=\linewidth]{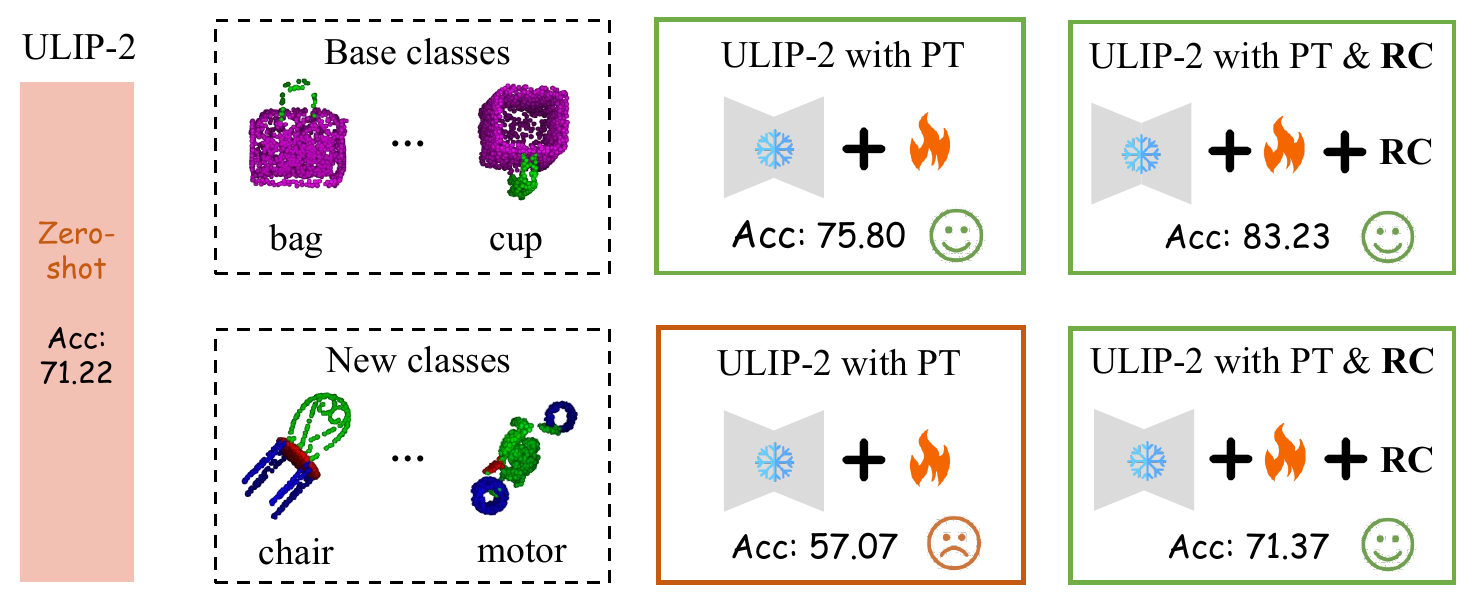}%
    \caption{\textbf{Motivation of our research: to promote the performances on downstream 3D tasks while maintaining good 
    generalization of large 3D models.} The experiments are conducted on ShapeNetCoreV2. ULIP-2 can reach 71.22\% zero-shot 
    recognition accuracy on this dataset. Recent works built on ULIP-2 introduce lightweight prompt tuning (PT) to further 
    boost target tasks (75.80\% accuracy). However, we observe the improvements come at the expenses of a severe drop in 
    3D domain generalization (e.g., 57.07\% accuracy on new classes, much behind 71.22\%), and 
    develop a systematic regulation constraint (\textbf{RC}) framework to address this challenge. }
    \label{fig:motivation}
\end{wrapfigure}
transferability is still suppressed compared to 
prevalent large 3D foundation models~\cite{zhang22pointclip,zhu23pointclip2,xue23ulip,zhang23i2pmae,zhang23cafo,zhang23parameter,xue23ulip2}, 
which have been pre-trained on numerous volume of 3D data~\cite{Objaverse,Objaversexl} and demonstrated promising zero-shot 
capability. 
Recent works stand on the shoulder of large 3D foundation models and push the boundary of downstream 3d tasks  
by parameter-efficient adaptation, such as prompt learning~\cite{zha23idpt,sun24ppt3d}, 
adapter~\cite{tang24pointpeft,zhou24dapt}, and their combination. 
They insert learnable prompts in the inputs or adapter inside the Transformer~\cite{vaswani17transformer} blocks to adapt the foundation models to specific 3D tasks. 
However, optimizing the newly introduced small modules targeting downstream benchmarks is prone to overfitting, 
thus disturbing the internal representations and compromising the inherent generalization of the foundation models~\cite{zhou22cocoop,lu22prompt,zhu23prograd,lee23rpo,khattak23maple,khattak23promptsrc}. 
As Fig.~\ref{fig:motivation} demonstrates, 
lightweight prompt tuning can notably lift the recognition accuracy of representative large 3D models on seen classes 
while hindering the generalization on unseen new classes, where the performances consistently lag behind 
corresponding zero-shot predictions of these models. 

In this paper, we develop our approach based on large 3D foundation models through lightweight prompt learning and 
propose a comprehensive framework that consists of three regulation constraints to allow the learning trajectory to 
interact with the well-learned knowledge in large 3D models actively, 
achieving better task-specific performances and task-agnostic generalization at the same time.
Specifically, we propose the mutual agreement constraint to regulate the learnable prompts to produce consistent feature distributions 
and predictions with the pre-trained foundation models. 
Then, we exploit the flexible and diverse text descriptions derived from LLMs or manual templates to reflect the attributes of different 
classes of point clouds and enhance the generalization. 
Finally, we develop a weighted model ensemble strategy to update the learnable prompts smoothly and predictably, 
avoiding giant and unexpected leaps toward overfitting the downstream datasets. 
Some recent works also explore parameter-efficient tuning for point cloud analysis~\cite{zha23idpt,tang24pointpeft,sun24ppt3d,zhou24dapt}, they focus on the performances of downstream tasks while failing to take the model generalization into account. 
As far as we are aware, our work initiates the first attempt to impose explicit regulation constraints and improve the 3D domain 
generalization based on large 3D models. 

In addition, we argue existing 3D domain generalization evaluation benchmarks, such as PointDA~\cite{qin19pointdan} and Sim-to-Real~\cite{huang21metasets},
may not be comprehensive to evaluate common generalization capabilities. 
Only $\sim$10 point cloud object classes are included in these benchmarks. %
They emphasize the generalization among the shared categories between the source and target domain, 
without considering transferring to unseen new classes, 
corrupted data, \textit{etc}, which are frequent in real-world scenarios. 
In this paper, 
three new benchmarks are created to enrich 3D domain generalization evaluation, including \emph{base-to-new class} generalization,
\emph{cross-dataset} generalization and \emph{few-shot} generalization. 
We will dissect the details of benchmark curation and usage in Section~\ref{subsec:new_3ddg_benchmarks}. %
We supply comprehensive experiments and analysis to examine the proposed regulation constraint framework, 
ablate the effectiveness of distinct components, and draw some new insights from our newly introduced 3DDG evaluation benchmarks. 
The results verify the proposed method not only enhances the task-specific 3D point cloud recognition but also extends the 
task-agnostic generalization ability by a clear margin. 

In short, the contributions of this work are threefold. 
\emph{\textbf{Firstly}}, to our knowledge, we firstly bring the 3DDG problem in front of large multi-modal 3D models and present an effective regulation framework based on lightweight prompt tuning, which not only strengthens downstream 3D task performances but also lifts the domain generalization capability remarkably. 
\emph{\textbf{Secondly}}, we implement our regulation framework as a plug-and-play module to seamlessly integrate into the existing large multi-modal 3D models. Consistent improvements are obtained over representative large 3D models, indicating the proposed regulation framework is general and model-agnostic. %
\emph{\textbf{Thirdly}}, we carefully craft three new benchmarks to enrich the evaluation of 3D domain generalization. Our benchmarks introduce new evaluation dimensions for 3DDG which are vital in the real world but absent in existing ones, including base-to-new, cross-dataset, and few-shot generalization. These new and more challenging benchmarks will drive the future research of 3D domain generalization.

\section{Related Work}

\textbf{3D domain generalization}. 
Although domain generalization has been widely studied in image recognition~\cite{ghifary15dg,li17dg,li18deepdg,li18dg_metalearning,li18dg_adversarial,li19episodic,li20dg_medical,zhao20dg,huang20self,zhou21dg_mixstyle,bui21exploiting,mahajan21dg,ding22dg,zhou23dg,zhang23dg}, 
it is still not the case for 3D. A large body of works in point cloud recognition focuses on improving the performances 
on specific benchmarks by supervised~\cite{qi17pointnet,qi17pointnet2,li18pointcnn,wu19pointconv,thomas19kpconv,wang19dgcnn,guo2020pct,xiang21curvenet,zhao21pt,ma22pointmlp,wu22ptv2,qian22pointnext,duan23condaformer,park23spotr} 
or self-supervised~\cite{yu22pointbert,pang22pointmae,zhang22pointm2ae,sun23vipformer} learning. 
However, they lack systematic strategies to address the generalization challenge and related evaluation is absent. 
Only a few methods investigate the 3D domain adaptation~\cite{qin19pointdan} and domain generalization~\cite{huang21metasets,wei22pdg} problem. 
They either create a common feature space between the source and target domain (e.g., PointDAN~\cite{qin19pointdan}, PDG~\cite{wei22pdg}), or 
utilize the meta-learning framework (e.g., MetaSets~\cite{huang21metasets}) to obtain robust representations to handle the domain shifts. 
Nevertheless, those methods are all built on small-size point cloud encoders targeting small-scale datasets and the overall generalization is unsatisfactory. 
In contrast, we explore the 3D domain generalization based on representative large multi-modal 3D models, like PointCLIP 
series~\cite{zhang22pointclip,zhu23pointclip2} and ULIP series~\cite{xue23ulip,xue23ulip2}. Meanwhile, we do not touch the 
backbone and only conduct lightweight prompt tuning on those large 3D models. 

\textbf{Prompt learning for large 3D models}. 
Prompt learning for 3D point cloud understanding has been studied in recent works~\cite{zha23idpt,sun24ppt3d,tang24pointpeft,zhou24dapt}. 
IDPT~\cite{zha23idpt}, Point-PEFT~\cite{tang24pointpeft} and DAPT~\cite{zhou24dapt} explore this problem in pure point cloud modality and do not establish 
connections with flexible language descriptions, thus these methods cannot conduct open-vocabulary 3D recognition. 
PPT~\cite{sun24ppt3d} firstly constructs a prompt learning pipeline based on the 
multi-modal framework ULIP~\cite{xue23ulip}. 
It achieves open-vocabulary recognition with promising performances and relatively small costs. 
Our work is closely related to PPT but distinguished in the following aspects: First, PPT focuses on optimizing specific 
3D tasks with learnable prompts and fails to consider generalization on unseen data. Instead, our work develops systematic
strategies to regulate prompt learning to boost generalization as well as target tasks. 
Second, PPT only introduces learnable prompts in the text branch while our method conducts multi-modal prompt tuning on 
both text and 3D branches. 

\section{Method}
\label{sec:method}

We firstly revisit lightweight prompt learning for existing large 3D models in section~\ref{subsec:mllm_prompt_tuning}. 
Then, a comprehensive regulation framework is proposed to promote the generalization capability of large 3D models
based on the plug-and-play prompt tuning strategy in section~\ref{subsec:regulated_prompt_tuning_framework}.
Finally, we introduce the implementation details of the devised method in section~\ref{subsec:impl_details}. 
The overall pipeline of our method is visualized in Fig.~\ref{fig:architecture}. 
The creation and analysis of our new 3DDG benchmarks are 
elaborated in Appendix~\ref{subsec:new_3ddg_benchmarks} due to space limitation. 

\subsection{Preliminary}
\label{subsec:mllm_prompt_tuning}
Existing large multi-modal 3D models~\cite{zhang22pointclip,zhu23pointclip2,xue23ulip,xue23ulip2} have different branches that encode the inputs from point cloud and text. 
In the 3D branch, a point cloud $P \in \mathbb{R}^{N\times 3}$ is divided and projected into $u$ point patches. 
Then, a class token $p_{cls} \in \mathbb{R}^{d}$ is inserted before the patches to form 
the input $\textbf{\textit{P}} = \{p_{cls}, p_1, p_2, \dots, p_u\} \in \mathbb{R}^{(1+u)\times d}$ of the 3D encoder $f_{P}(\cdot, \theta_{p})$,
where $\theta_p$ represents the encoder parameters.
In the text branch, the descriptions of each 3D category  
are converted into the sequence $\textbf{\textit{T}} = \{t_{sos}, t_1, t_2, \dots, t_v, t_c, t_{eos}\} \in \mathbb{R}^{(3+v)\times d}$ for 
the text encoder $f_{T}(\cdot, \theta_{t})$. Here $t_c$ is the embedding of \verb|{class}|, 
$t_{sos}$ and $t_{eos}$ stands for the the start and end flag token of a sentence. 
So we can obtain the 3D features $\textbf{h}_{P} = f_{P}(\textbf{\textit{P}}, \theta_p)$ and text features $\textbf{h}_{T} = f_{T}(\textbf{\textit{T}}, \theta_t)$. 
When executing zero-shot recognition for a downstream 3D dataset of $C$ categories, 
$\theta_p$ and $\theta_t$ are frozen, the model outputs the class probability distribution $\mathcal{D}$ for point cloud $P$ 
by computing \begin{math}\frac{\exp({sim(\textbf{h}_P,\ \textbf{h}_T)}/\tau)}{\sum_{j=1}^{C} \exp({sim(\textbf{h}_P,\ \textbf{h}_T^j)}/\tau)}\end{math},
where $sim(\cdot, \cdot)$ measures the cosine similarity of the inputs and $\tau$ is a temperature coefficient.

Although zero-shot inference is flexible, the performances on target tasks may not be satisfactory. 
Multi-modal prompt learning introduces learnable prompts in the inputs of different branches. Specifically, 
we insert $r$ learnable prompts $\textbf{\textit{E}}^P = \{e_1^P, e_2^P, \dots, e_r^P\} \in \mathbb{R}^{r\times d}$ into $\textbf{\textit{P}}$
and $s$ learnable prompts $\textbf{\textit{E}}^T = \{e_1^T, e_2^T, \dots, e_s^T\} \in \mathbb{R}^{s\times d}$ into $\textbf{\textit{T}}$, respectively. 
Thereupon, the modified inputs for point cloud and text encoder become 
$\tilde{\textbf{\textit{P}}} = \{p_{cls}, p_1, \dots, p_u, e_1^P, \dots, e_r^P\}$
and $\tilde{\textbf{\textit{T}}} = \{t_{sos}, t_1, t_2, \dots, t_v, t_c, e_1^T, \dots, e_s^T, t_{eos}\}$. 
After transforming by the encoders, we obtain the new point cloud and text representations 
denoted with $\tilde{\textbf{h}}_{P} = f_P(\tilde{\textbf{\textit{P}}}, \tilde{\theta}_p)$
and $\tilde{\textbf{h}}_{T} = f_T(\tilde{\textbf{\textit{T}}}, \tilde{\theta}_t)$, where 
$\tilde{\theta}_p = \{\theta_p, \textbf{\textit{E}}^P\}$ and $\tilde{\theta}_t = \{\theta_t, \textbf{\textit{E}}^T\}$. 
Similarly, the predicted class distribution $\tilde{\mathcal{D}}$ and optimization objective can be formulated by Eq.~\ref{eq:ulip_multimodal_pl}. 
\begin{equation}
   \tilde{\mathcal{D}} = \frac{\exp(sim(\tilde{\textbf{h}}_{P}, \tilde{\textbf{h}}_{T}) / \tau)}
   {\sum_{j=1}^{C} \exp(sim(\tilde{\textbf{h}}_{P}, \tilde{\textbf{h}}_{T}^j) / \tau)}, \qquad \{\textbf{\textit{E}}^{P*},\ \textbf{\textit{E}}^{T*}\} = \argmin_{\{\textbf{\textit{E}}^P,\ \textbf{\textit{E}}^T\}} \mathbb{E}_{(P, y)\sim \mathcal{D}_{gt}} \mathcal{L}_{CE}(\tilde{\mathcal{D}}, y)
   \label{eq:ulip_multimodal_pl}
\end{equation}
where $\mathcal{D}_{gt}$ is the ground truth distribution of point cloud data and $y$ is the category of point cloud $P$. 
Note that $\theta_p$ and $\theta_t$ are still frozen and only $\textit{E}^P$ and $\textit{E}^T$ are updatable with
the cross entropy ($CE$) loss. 
Also, the learnable prompts can be inserted at each layer of the 3D and text encoder, not only at the very first layer. We call this scheme deep multi-modal prompt learning that will be regarded as an important baseline in our experiment settings. 

\begin{figure}[t]
    \centering
    \includegraphics[width=\linewidth]{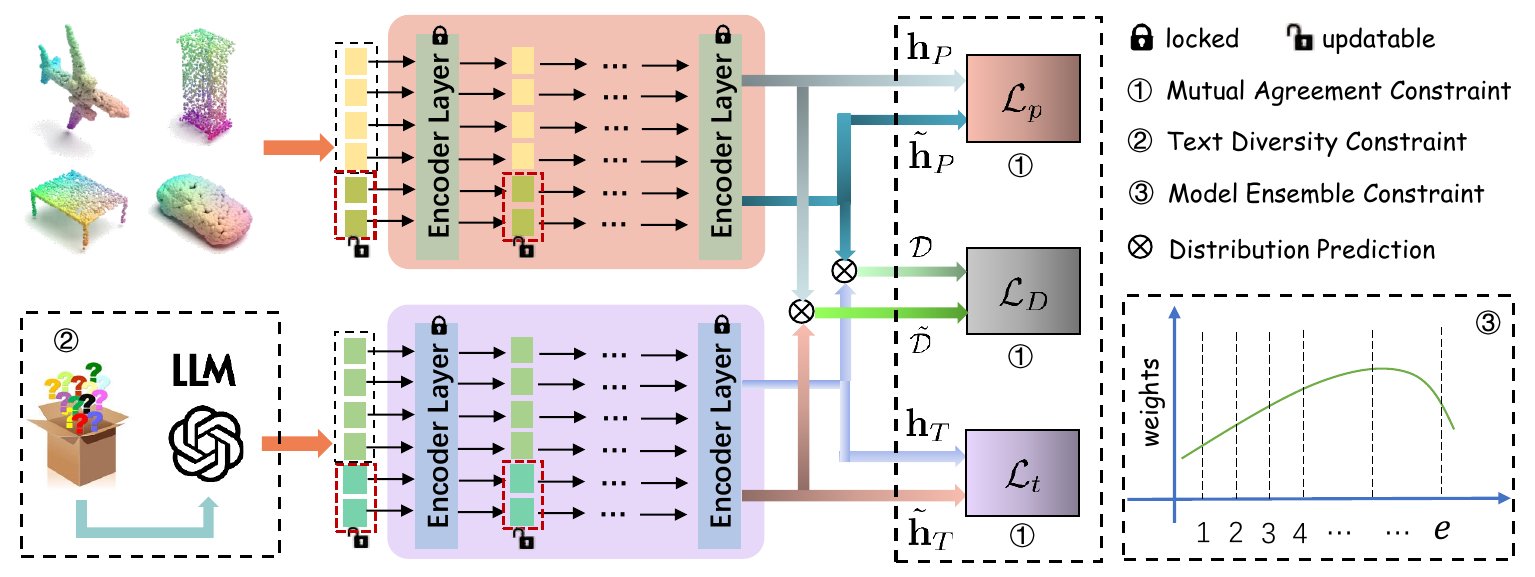}
    \caption{\textbf{The overall architecture of our point cloud analysis prompt regulation constraint framework}, namely \textbf{Point-PRC}, consisting of three core components as in the figure.}
    \label{fig:architecture}
\end{figure}

\subsection{Our Regulation Framework}
\label{subsec:regulated_prompt_tuning_framework}
Prompt learning aims to elicit well-learned knowledge of pre-trained large models by introducing a small number of
learnable parameters in the input space. But optimizing the learnable prompts targeting specific datasets easily compromises the general knowledge. 
To handle the above problems, we propose a comprehensive regulation framework consisting of 
three components: mutual agreement constraint, text diversity constraint, and 
model ensemble constraint, as elaborated below. 

\subsubsection{Mutual Agreement Constraint (MAC)}%

Large foundation models unfold overall better 
robustness and transferability on a broad spectrum of evaluations than conventional models learned on specific datasets, 
supported by representative works in vision~\cite{dehghani23scaling}, language~\cite{radford2019gpt2,raffel20t5,brown20gpt3}, and 
multi-modal understanding~\cite{radford21clip,jia21scaling,fang22data,pham23combined,zhou22coop,zhou22cocoop,zhu23pointclip2,xue23ulip,guo23pointbind}. 
The first component of the proposed framework is to interact with large 3D models actively
by maximizing the mutual agreement between learnable prompts and pre-trained knowledge. 

Specifically, we engage with large 3D models by aligning extracted features and predicted distributions simultaneously. 
Let us denote the frozen point cloud feature extracted by the 3D foundation model as $\textbf{h}_{P}$, 
the point cloud feature containing learnable prompts as $\tilde{\textbf{h}}_{P}$.
Now we compute the difference between $\textbf{h}_{P}$ and $\tilde{\textbf{h}}_{P}$ and mark it as $\mathcal{L}_{p}$. 
Similarly, in the text modality, we have $\mathcal{L}_{t}$ which measures the difference between $\textbf{h}_{T}$ and $\tilde{\textbf{h}}_{T}$. 
On the other side, $\mathcal{D}$ and $\tilde{\mathcal{D}}$ are two class distributions given by 
the frozen and promptable large 3D models, respectively. The difference between $\mathcal{D}$ and $\tilde{\mathcal{D}}$ is 
denoted as $\mathcal{L}_{D}$. 
Our mutual agreement constraint aims to minimize the feature and prediction distribution discrepancy to 
ensure the learning trajectory not to forget the task-agnostic knowledge in large pre-trained models.
\begin{equation}
    \mathcal{L}_{p} = \sum_{i} |\ \textbf{h}^{i}_{P} - \tilde{\textbf{h}}^{i}_{P}\ |, \quad 
    \mathcal{L}_{t} = \sum_{i} |\ \textbf{h}^{i}_{T} - \tilde{\textbf{h}}^{i}_{T}\ |, \quad
    \mathcal{L}_{D} = \sum_{i} \mathcal{D}_{KL}(\mathcal{D}_i\ ||\ \tilde{\mathcal{D}}_i)
    \label{eq:mutual_agreement_constraint}
\end{equation}
As formulated in Eq.~\ref{eq:mutual_agreement_constraint}, L$_1$ distance is employed to compute $\mathcal{L}_{p}$
and $\mathcal{L}_{t}$, and Kullback-Leibler (KL) divergence is used to characterize the distribution discrepancy. 
We will examine the design choices in the ablation study. 

\subsubsection{Text Diversity Constraint (TDC)}
\begin{figure}[ht]
    \centering
    \includegraphics[width=\linewidth]{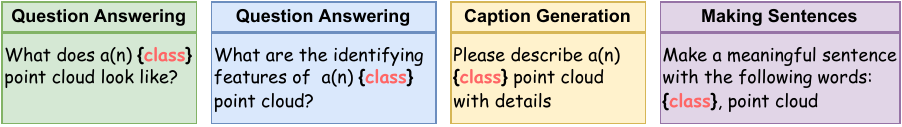}
    \caption{\textbf{Illustration of diverse questions to LLMs}, including GPT-3.5, GPT-4 and PointLLM. 
    The responses given by LLMs are regarded as the text descriptions to the point cloud and fed into the text encoder.}
    \label{fig:regulation_constraint_tdc}
\end{figure}

Inspired by the flexibility and versatility of language expressions, we propose to leverage diverse text descriptions to guide 
the lightweight prompt tuning to produce transferrable features. 
Specifically, we obtain multiple text descriptions for each point cloud object category by prompting LLMs 
($e.g.$, GPT-3.5~\cite{chatgpt}, GPT-4~\cite{gpt4weburl}, PointLLM~\cite{xu23pointllm}) or utilizing manual templates. 
Then, we aggregate the text feature of all descriptions for each single category by pooling operation, 
$\textbf{h}_T = \textrm{AvgPool}(\sum_j \textbf{h}_T^j)$, which will 
integrate rich semantic information extracted by powerful large models, prevent a point cloud category biasing 
towards some specific descriptions and finally enhance the model transferability. 
In the case of describing point clouds with LLMs, 
we design three kinds of prompts, including question answering, caption generation, and making sentences using keywords, as demonstrated in 
Fig.~\ref{fig:regulation_constraint_tdc}. 
For each instruction to the LLM, we acquire $N_t = 10$ responses. 

\subsubsection{Model Ensemble Constraint (MEC)}

The model ensemble constraint aims to synthesize the opinions from different models by weighted voting 
to avoid some extreme and failure cases of a single model. The idea has been widely discussed in statistical machine 
learning~\cite{opitz99ensemble,dietterich00ensemble} and deep learning~\cite{ganaie22ensemble,mohammed23ensemble}. 
Robust tuning of multi-modal large models by ensemble learning also has been studied in recent 
literature~\cite{wortsman22wiseft,ilharco22paint}. The ensemble strategy mainly involves interpolating weights between 
zero-shot and fully fine-tuned large models. 
But it has not been investigated in the context of prompt tuning for large 3D models and its effectiveness is unknown. 
In this paper, we propose to ensemble models by aggregating the model parameters in different training epochs 
with a Gaussian weighted strategy. 
The basic idea is that in the initial learning stage, the prompts are randomly initialized and 
not well optimized so we distribute them very small weights. As the training iterates, the model gradually gets a 
sense of downstream tasks; thus, increasing weights are assigned to the model parameters in these epochs. 
As the training ends, the learnable prompts are adjusted well to adapt downstream datasets while having 
the risk of overfitting, so we decrease the weights to the model parameters. 
The varying weights of the above process can be approximated by a gaussian curve. 
Finally, the weighted models in different epochs are ensembled to generate the model parameters 
$\tilde{\theta}_t$ and $\tilde{\theta}_p$, shown in Eq.~\ref{eq:model_ensemble_constraint}.
\begin{equation}
    \tilde{\theta}_p = \sum_{i=1}^e w_i \tilde{\theta}_p^i, \quad \tilde{\theta}_t = \sum_{i=1}^e w_i \tilde{\theta}_t^i
   \label{eq:model_ensemble_constraint}
\end{equation}
where $e$ is the number of epochs and 
$w_i = \frac{1}{\sigma \sqrt{2\pi}} \exp(-\frac{(i-\mu)^2}{2\sigma^2})$. $\mu$ and $\sigma^2$ represent the mean 
and variance of a gaussian distribution. 
$\tilde{\theta}_p^i = \{\theta_t^i, E^{P_i}\}$ and $\tilde{\theta}_t^i = \{\theta_t^i, E^{T_i}\}$ indicate the model 
parameters after the $i$th epoch of training in the text and point cloud branch, respectively. 
Note that a simple accumulated addition can implement Eq.~\ref{eq:model_ensemble_constraint} and 
we do not need to store all $e$ copies of the parameters, referring to Appendix for details. 

\textbf{Optimization.}
The overall optimization objective consists of two parts, the task-specific cross entropy loss $\mathcal{L}_{CE}$ and 
the task-agnostic regulation constraint loss $\mathcal{L}_{RC}$, displayed in Eq.~\ref{eq:overall_loss},
where $\alpha, \beta, \gamma$ are hyperparameters. 
Unlike trivial prompt tuning a multi-modal large model on downstream tasks, this design allows the learnable prompts to 
actively interact and align with the general knowledge in a pre-trained large model while learning on specific 3D tasks.
\begin{equation}
    \mathcal{L} = \mathcal{L}_{CE} + \mathcal{L}_{RC},\qquad \mathcal{L}_{RC} = \alpha\mathcal{L}_p + \beta\mathcal{L}_t + \gamma\mathcal{L}_{D}
    \label{eq:overall_loss}
\end{equation}

\subsection{Implementation Details}
\label{subsec:impl_details}
We choose PointCLIP~\cite{xue23ulip}, PointCLIP V2~\cite{xue23ulip}, ULIP~\cite{xue23ulip}, and ULIP-2~\cite{xue23ulip2} as the 3D foundation models for experiments. 
All experiments are running with three random seeds and we report the mean and standard deviation. 
The learnable prompts are inserted into the inputs of first 9 Transformer layers in these models and 
the prompt length is set to 2. 
Unless specified, the prompts are optimized using 16-shot learning. 
Note that previous 3DDG methods~\cite{qin19pointdan,huang21metasets,wei22pdg} use the full training set. 
We set $\alpha = 10$, $\beta = 25$ and $\gamma = 1$.
The optimizer is SGD, the initial lr is 0.0025 and we use cosine scheduler to update it. 
More details about the model configuration can be found in Appendix.  
We will justify the design choices in the ablation study. 

\section{Experiments}
\label{sec:experiments}
In this section, we first explain the evaluation settings of our newly curated and existing benchmarks. 
Then, comprehensive comparison and analysis across various generalization settings are presented to 
show the advantages of the proposed method. Finally, we justify the effectiveness of different components 
in our regulation framework through systematic controlled experiments. 

\subsection{3DDG Evaluation Settings}
\textbf{Base-to-New}. This benchmark includes 5 point cloud datasets which are ModelNet40~\cite{modelnet}, three variants of ScanObjectNN~\cite{uy19sonn} (S-PB\_T50\_RS, S-OBJ\_BG, S-OBJ\_ONLY)
and ShapeNetCoreV2~\cite{shapenet2015}. 
Each dataset is equally split into base and new classes, where the former is used for prompt tuning while the latter only serves the test purpose. 
\textbf{Cross-Dataset}. This benchmark has four types of evaluation, including \emph{OOD generalization}, \emph{data corruption}, \emph{PointDA}~\cite{qin19pointdan} and \emph{Sim-to-Real}~\cite{huang21metasets}.
We established the first two assessments and the latter two already existed. 
For \emph{OOD generalization}, models are trained on the source domain and evaluated on the target domains.  
For \emph{data corruption}, models are trained on clean ModelNet~\cite{modelnet} and tested on the corrupted data 
in ModelNet-C\cite{ren22modelnet-c}. 
\textbf{Few-Shot}. This setting inspects the model generalization in an extremely low-data regime, where 1, 2, 4, 8, and 16 shots are randomly sampled for prompt learning, and 
the recognition accuracy is calculated on the whole test set, respectively. The explanations are brief and we encourage the readers to check the details in Appendix. 

\subsection{Base-to-new Class Generalization}

In this benchmark, models are learned on the base classes and evaluated on the test sets of base and novel classes. 
In addition to ULIP and ULIP-2, we also implement the same prompt tuning for PointCLIP~\cite{zhang22pointclip} (P-CLIP) and 
PointCLIP V2~\cite{zhu23pointclip2} (P-CLIP2) for comparison, 
shown in Tab.~\ref{tab:base2new_generalization}.

\textbf{Loss of Generalization in P-CLIP and ULIP Series}. We observe notable gaps occur between base and new class 
recognition accuracy of P-CLIP, P-CLIP2, ULIP, ULIP-2 when prompt tuning without the proposed regulation constraints.  
For instance, P-CLIP2 achieves 93.98\% accuracy on the base classes of ModelNet40 while dropping by 48.77\%
absolute points on the whole test set of the new classes, which even lags behind the zero-shot accuracy of the frozen 
P-CLIP2 (64.22\%). The results are consistent across five datasets, suggesting the loss of generalization of original models.  

\textbf{Lifting the Generalization by Our Framework}. As shown in Tab.~\ref{tab:base2new_generalization}, the proposed framework composite of three regulation constraints boosts the unseen 
class recognition accuracy across different models and datasets by a clear margin, thanks to the active communication 
and alignment with the general knowledge in large 3D models. For example, the improvement of the harmonic mean on ULIP 
reaches 10.65\% absolute points averaged over 5 datasets. 

\textbf{Lifting the Specific 3D Tasks by Our Framework}. Surprisingly, the task-specific performances are not be hindered 
by the regulation constraints while enhancing the task-agnostic generalization, referring to the base class accuracy of 
ULIP+\textbf{RC} and ULIP-2+\textbf{RC} averaged over 5 datasets, increasing by 4.87\% and 5.27\%, respectively. 

\begin{table*}[t]\tiny
   \centering
   \caption{\textbf{Base-to-new class generalization comparison for representative large 3D models based on prompt learning}. 
   Each number here is the mean of three runnings. 
   Base: base class accuracy (in \%, same below). New: new class accuracy. HM: harmonic mean of base and new class accuracy. 
   +\textbf{RC} demonstrates the models with our regulation constraint framework. 
   }
   \begin{subtable}{0.32\linewidth}
      \caption{\textbf{Average over 5 datasets}}
      \begin{tabular}{l c c | c}
      \toprule
      Method & Base & New & HM \\
      \midrule
      P-CLIP~\cite{zhang22pointclip} & 75.66 & 23.45 & 35.80 \\
      P-CLIP2~\cite{zhu23pointclip2} & 74.11 & 37.84 & 50.10 \\ %
      \midrule
      ULIP~\cite{xue23ulip} & 77.32 & 49.01 & 59.99 \\
      +\textbf{RC}(Ours) & \textbf{82.19} & \textbf{61.93} & \textbf{70.64} \\
      \midrule
      ULIP-2~\cite{xue23ulip2} & 77.91 & 67.91 & 72.57 \\ %
      +\textbf{RC}(Ours) & \textbf{83.18} & \textbf{76.10} & \textbf{79.48} \\
      \bottomrule
      \end{tabular}
      \label{tab:base2new_avg_five_datasets}
   \end{subtable}
   \quad
   \begin{subtable}{0.32\linewidth}
      \caption{ModelNet40}
      \begin{tabular}{l c c | c}
      \toprule
      Method & Base & New & HM \\
      \midrule
      P-CLIP~\cite{zhang22pointclip} & 93.23 & 20.22 & 33.23\\
      P-CLIP2~\cite{zhu23pointclip2} & 93.98 & 45.21 & 61.05\\
      \midrule
      ULIP~\cite{xue23ulip} & 92.80 & 50.07 & 65.05 \\
      +\textbf{RC}(Ours) & \textbf{95.03} & \textbf{55.27} & \textbf{69.89} \\
      \midrule
      ULIP-2~\cite{xue23ulip2} & 91.77 & 56.47 & 69.92 \\ %
      +\textbf{RC}(Ours) & \textbf{95.30} & \textbf{64.83} & \textbf{77.17}\\
      \bottomrule
      \end{tabular}
      \label{tab:base2new_mn40}
   \end{subtable}
   \quad
   \begin{subtable}{0.32\linewidth}
      \caption{S-PB\_T50\_RS}
      \begin{tabular}{l c c | c}
      \toprule
      Method & Base & New & HM \\
      \midrule
      P-CLIP~\cite{zhang22pointclip} & 61.25 & 19.87 & 30.01\\
      P-CLIP2~\cite{zhu23pointclip2} & 56.84 & 29.92 & 39.20\\
      \midrule
      ULIP~\cite{xue23ulip} & 56.73 & 25.80 & 35.47 \\
      +\textbf{RC}(Ours) & \textbf{64.20} & \textbf{49.17} & \textbf{55.69} \\
      \midrule
      ULIP-2~\cite{xue23ulip2} & 66.40 & 66.47 & 66.43 \\
      +\textbf{RC}(Ours) & \textbf{73.67} & \textbf{74.27} & \textbf{73.97}\\
      \bottomrule
      \end{tabular}
      \label{tab:base2new_so_pb_t50_rs}
   \end{subtable}
   \begin{subtable}{0.32\linewidth}
      \vspace{9pt}
      \caption{S-OBJ\_BG}
      \begin{tabular}{l c c | c}
      \toprule
      Method & Base & New & HM \\
      \midrule
      P-CLIP~\cite{zhang22pointclip} & 72.82 & 23.00 & 34.96 \\
      P-CLIP2~\cite{zhu23pointclip2} & 70.07 & 35.08 & 46.75\\
      \midrule
      ULIP~\cite{xue23ulip} & 73.20 & 47.17 & 57.37 \\
      +\textbf{RC}(Ours) & \textbf{79.47} & \textbf{55.20} & \textbf{65.15} \\
      \midrule
      ULIP-2~\cite{xue23ulip2} & 77.00 & 83.27 & 80.01 \\
      +\textbf{RC}(Ours) & \textbf{80.10} & \textbf{88.93} & \textbf{84.28}\\
      \bottomrule
      \end{tabular}
      \label{tab:base2new_so_obj_bg}
   \end{subtable}
   \quad
   \begin{subtable}{0.32\linewidth}
      \vspace{9pt}
      \caption{S-OBJ\_ONLY}
      \begin{tabular}{l c c | c}
      \toprule
      Method & Base & New & HM \\
      \midrule
      P-CLIP~\cite{zhang22pointclip} & 76.23 & 20.23 & 31.97\\
      P-CLIP2~\cite{zhu23pointclip2} & 71.40 & 44.39 & 54.74\\
      \midrule
      ULIP~\cite{xue23ulip} & 74.13 & 50.80 & 60.29 \\
      +\textbf{RC}(Ours) & \textbf{79.23} & \textbf{65.93} & \textbf{71.97} \\
      \midrule
      ULIP-2~\cite{xue23ulip2} & 78.60 & 76.27 & 77.42 \\
      +\textbf{RC}(Ours) & \textbf{83.60} & \textbf{81.10} & \textbf{82.33}\\
      \bottomrule
      \end{tabular}
      \label{tab:base2new_so_obj_only}
   \end{subtable}
   \quad
   \begin{subtable}{0.32\linewidth}
      \vspace{9pt}
      \caption{ShapeNetCoreV2}
      \begin{tabular}{l c c | c}
      \toprule
      Method & Base & New & HM \\
      \midrule
      P-CLIP~\cite{zhang22pointclip} & 74.78 & 33.92 & 46.61\\
      P-CLIP2~\cite{zhu23pointclip2} & 78.27 & 34.58 & 47.97\\
      \midrule
      ULIP~\cite{xue23ulip} & 89.73 & 71.20 & 79.40 \\
      +\textbf{RC}(Ours) & \textbf{93.03} & \textbf{84.10} & \textbf{88.34} \\
      \midrule
      ULIP-2~\cite{xue23ulip2} & 75.80 & 57.07 & 65.38 \\ %
      +\textbf{RC}(Ours) & \textbf{83.23} & \textbf{71.37} & \textbf{76.85}\\
      \bottomrule
      \end{tabular}
      \label{tab:base2new_snv2}
   \end{subtable}
   \label{tab:base2new_generalization}
\end{table*}

\subsection{Cross-Dataset Generalization}
This setting
differs from the base-to-new counterpart where the base and new classes belong to the same dataset. We present the analysis for \emph{OOD generalization} and \emph{data corruption} as below, and put the comparison on \emph{Sim-to-Real} and \emph{PointDA} in Appendix.

\emph{OOD Generalization} 
demonstrates the models' transferability to other unseen domains by learning from an existing domain. To evaluate on this benchmark, 
we implement the lightweight prompt learning for ULIP and ULIP-2 then impose the proposed regulation constraints on them. 
Prompt learning for P-CLIP~\cite{zhang22pointclip} and P-CLIP2~\cite{zhang22pointclip} with same settings are also implemented 
for comparison. 
The results are reported in Tab.~\ref{tab:xset_dg} . 
By wrapping ULIP and ULIP-2 with the devised framework, we achieve consistent positive gains on each of the 
five target domains. The average gains over them are enlarged with increasing ability of ULIP, $e.g.$, 
+6.20\% for ULIP-2 vs. +1.79\% for ULIP. Meanwhile, we notice that the performances on Omni3D~\cite{OmniObject3D} are rather 
limited and the methods here seem not to work, especially for P-CLIP series and ULIP (less than 10\% accuracy). 
This dataset contains a large vocabulary of real 3D objects (216 categories) and exhibits the long-tail attribute.
When transferring the models that learn from a narrow set of 3D object concepts (55 classes in ShapeNetV2)
to Omni3D, they suffer from new 3D concepts thus perform poorly. 

\begin{table*}
   \centering\scriptsize
   \caption{\textbf{Comparison of OOD generalization in cross-dataset benchmark}. ShapeNetV2 serves as the source domain and the other 
   five datasets are deployed as the target domain. ShapeNetV2: 55 classes, ModelNet40: 40 classes, SONN: 15 classes, Omni3D: 216 classes. Some common categories are shared between the source and target domain. Note that Omni3D has much more new 3D object concepts than others. The last column indicates the average over five target datasets.}
   \label{tab:xset_dg}
   \begin{tabular}{l c c c c c c c c}
      \toprule
      \multirow{2}{*}{Method} & \textbf{Source} & & \multicolumn{5}{c}{\textbf{Target}} & \multirow{2}{*}{\textbf{Avg.}} \\\cline{2-2}\cline{4-8}
             & ShapeNetV2 & & ModelNet40 & S-PB\_T50\_RS & S-OBJ\_BG & S-OBJ\_ONLY & Omni3D & \\
      \midrule
      P-CLIP~\cite{zhang22pointclip} & 67.41(0.09) & & 33.20(1.86) & 15.51(0.58) & 18.59(1.40) & 22.89(2.32) & 0.48(0.17) & 22.55(1.54) \\
      \midrule
      P-CLIP2~\cite{zhu23pointclip2} & 68.93(1.43) & & 54.73(1.48) & 39.53(4.22) & \textbf{34.30}(1.28) & \textbf{25.63}(1.16) & 8.63(2.52) & 32.56(2.13) \\
      +\textbf{RC}(Ours) & \textbf{69.80}(2.86) & & \textbf{55.37}(1.78) & \textbf{39.77}(0.45) & 34.20(0.54) & 24.50(1.26) & \textbf{10.20}(0.40) & \textbf{32.81}(0.89) \\
      \midrule
      ULIP~\cite{xue23ulip} & 87.33(0.95) & & 56.17(1.15) & 26.83(2.15) & 39.43(2.17) & 43.53(1.32) & 6.37(0.90) & 34.47(1.54) \\
      +\textbf{RC}(Ours) & \textbf{90.43}(0.86) & & \textbf{58.00}(0.57) & \textbf{28.43}(0.68) & \textbf{40.33}(0.71) & \textbf{46.33}(1.54) & \textbf{8.20}(0.50) & \textbf{36.26}(0.80) \\
      \midrule
      ULIP-2~\cite{xue23ulip2} & 76.70(1.37) & & 65.27(0.66) & 40.07(0.34) & 53.80(1.78) & 48.53(1.72) & 17.27(0.54) & 44.99(1.01) \\ 
      +\textbf{RC}(Ours) & \textbf{76.70}(1.59) & & \textbf{72.10}(0.93) & \textbf{46.77}(2.43) & \textbf{59.03}(3.02) & \textbf{56.27}(0.97) & \textbf{21.80}(0.49) & \textbf{51.19}(1.57) \\ %
      \bottomrule
   \end{tabular}
\end{table*}

\emph{Data Corruption} 
are common in point clouds due to complex geometry, sensor inaccuracy and processing imprecision. 
We investigate the generalization of the proposed framework on ModelNet-C~\cite{ren22modelnet-c}, which includes 
common corruptions, such as dropping some parts or adding global outliers. 
The compared methods are same as those in OOD generalization and the results are exhibited in 
Tab.~\ref{tab:xset_corruption_generalization}. 
Our method not only boosts the recognition accuracy on clean data (+1.44\% for ULIP and +1.40\% for ULIP-2), 
but also strengthen the robustness of representative large 3D models against collapsed data. 
By averaging on 7 types of corruption, we receive +1.51\% and +4.78\% gains for ULIP and ULIP-2, respectively. 
\begin{table*}\tiny
   \centering
   \caption{\textbf{Comparison of corruption generalization on ModelNet-C\cite{ren22modelnet-c} when trained on clean data}. 
   The results are reported for the corruption severity=2 in ModelNet-C.}
   \label{tab:xset_corruption_generalization}
   \begin{tabular}{l c c c c c c c c c}
      \toprule
      \multirow{2}{*}{Method} & \textbf{Clean Data} & \multicolumn{7}{c}{\textbf{Corruption Type}} & \multirow{2}{*}{\textbf{Avg.}} \\\cline{3-9}
             & ModelNet & Add Global & Add Local & Drop Global & Drop Local & Rotate & Scale & Jitter & \\
      \midrule
      P-CLIP~\cite{zhang22pointclip} & 80.97(1.02) & 80.97(1.02) & 80.97(1.02) & 64.95(1.08) & 68.31(1.93) & 65.75(1.19) & 72.04(1.33) & 52.09(1.28) & 69.30(1.26) \\
      P-CLIP2~\cite{zhu23pointclip2} & 83.49(0.51) & 83.49(0.51) & 83.49(0.51) & 68.85(3.22) & 66.67(1.96) & 70.13(1.33) & 75.68(0.15) & 61.21(2.16) & 72.79(1.41) \\ %
      \midrule
      ULIP~\cite{xue23ulip} & 82.43(1.25) & 82.50(0.99) & 82.27(1.17) & 80.77(1.03) & 65.43(1.02) & 72.27(1.56) & 74.67(1.58) & \textbf{45.60}(0.65) & 71.93(1.14) \\
      +\textbf{RC}(Ours) & \textbf{83.87}(0.34) & \textbf{83.83}(0.40) & \textbf{83.93}(0.19) & \textbf{81.83}(0.52) & \textbf{67.37}(1.72) & \textbf{79.10}(0.36) & \textbf{76.37}(0.09) & 41.67(4.79) & \textbf{73.44}(1.15) \\ %
      \midrule
      ULIP-2~\cite{xue23ulip2} & 85.07(0.21) & 81.97(0.79) & 82.03(0.96) & 79.93(0.92) & 60.03(1.21) & 80.30(0.93) & 75.77(0.74) & 44.27(2.13) & 72.04(1.10) \\ %
      +\textbf{RC}(Ours) & \textbf{86.47}(0.56) & \textbf{86.57}(0.48) & \textbf{86.30}(0.51) & \textbf{84.87}(0.48) & \textbf{67.80}(1.20) & \textbf{84.60}(0.22) & \textbf{81.17}(1.05) & \textbf{46.43}(2.45) & \textbf{76.82}(0.91) \\
      \bottomrule
   \end{tabular}
\end{table*}

\subsection{Few-shot Generalization}

\begin{figure*}[t]\scriptsize
   \centering
   \includegraphics[width=0.33\linewidth]{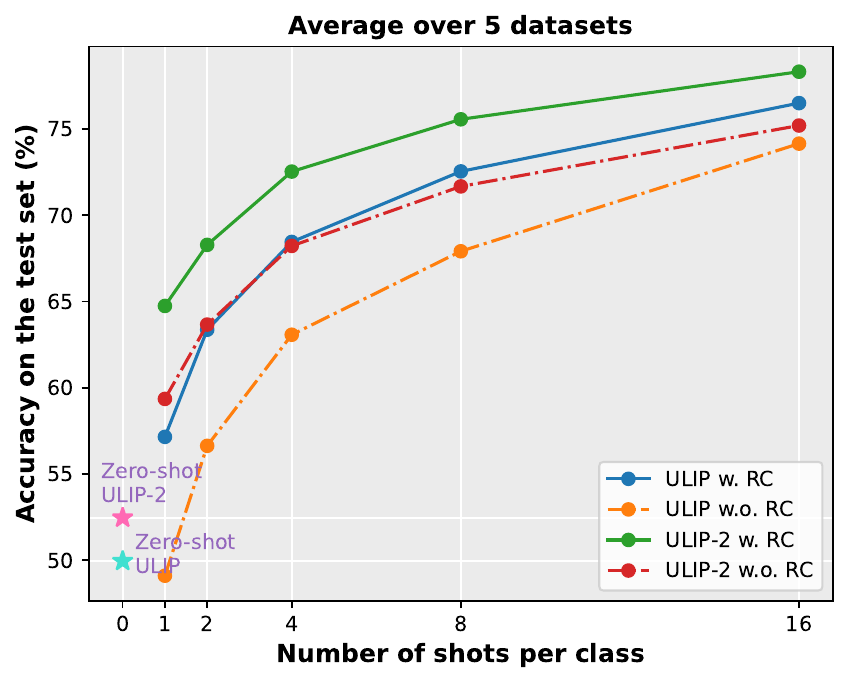}%
   \includegraphics[width=0.33\linewidth]{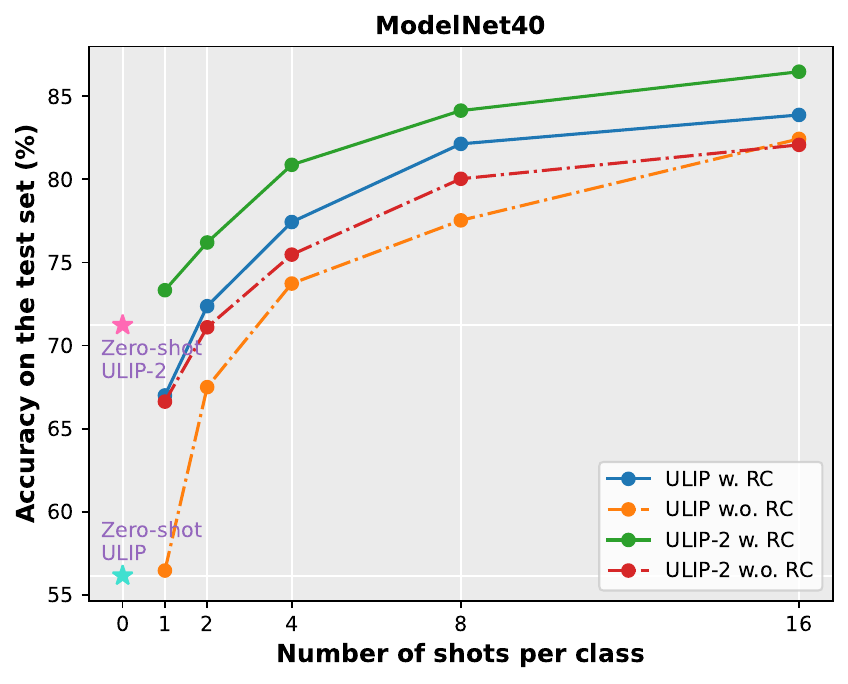}%
   \includegraphics[width=0.33\linewidth]{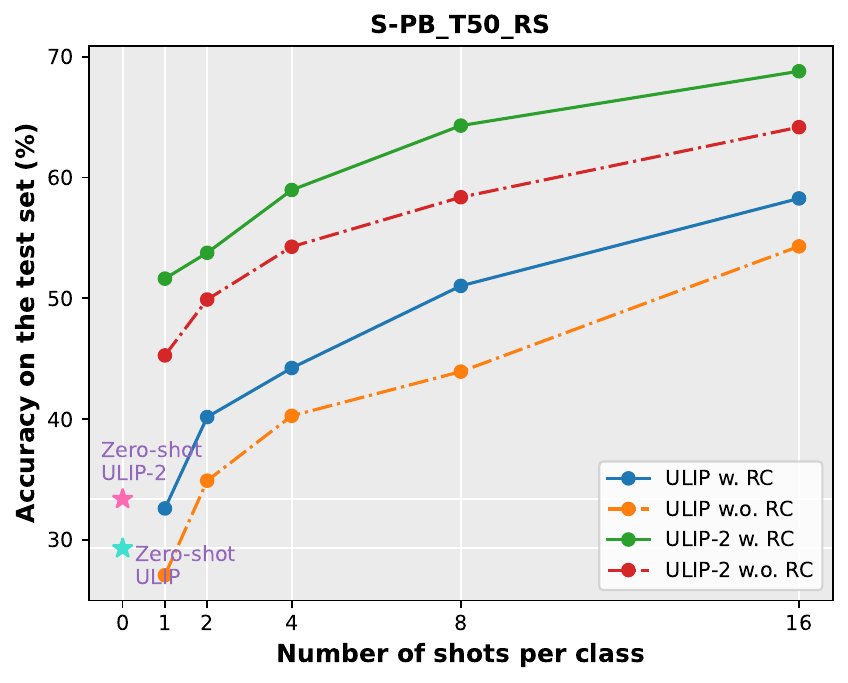}
   \includegraphics[width=0.33\linewidth]{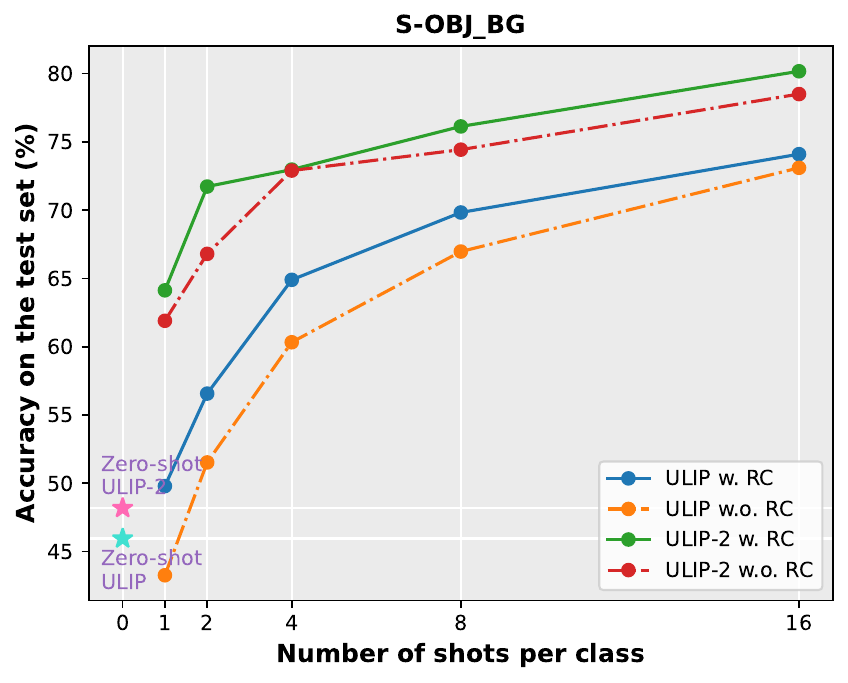}%
   \includegraphics[width=0.33\linewidth]{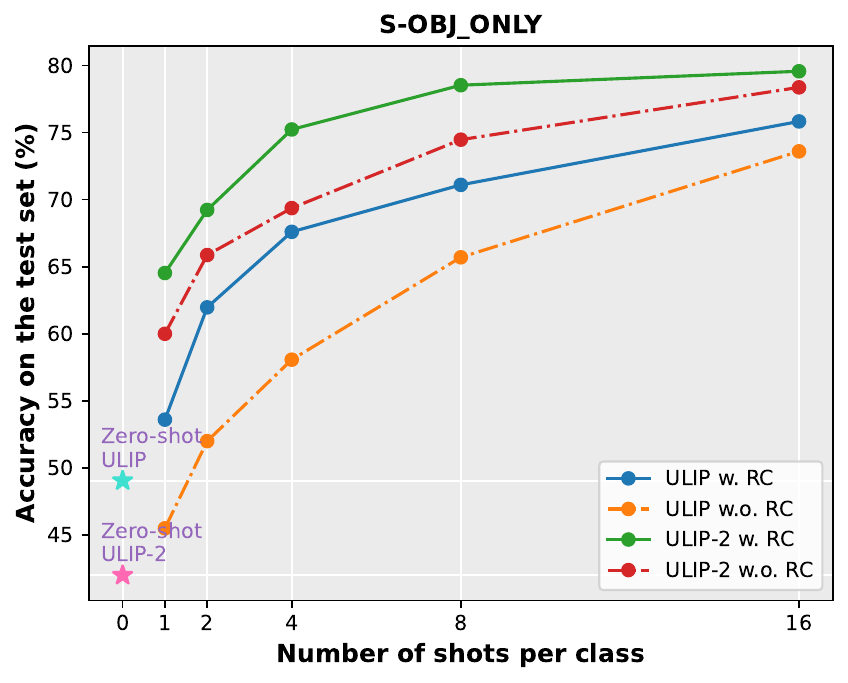}%
   \includegraphics[width=0.33\linewidth]{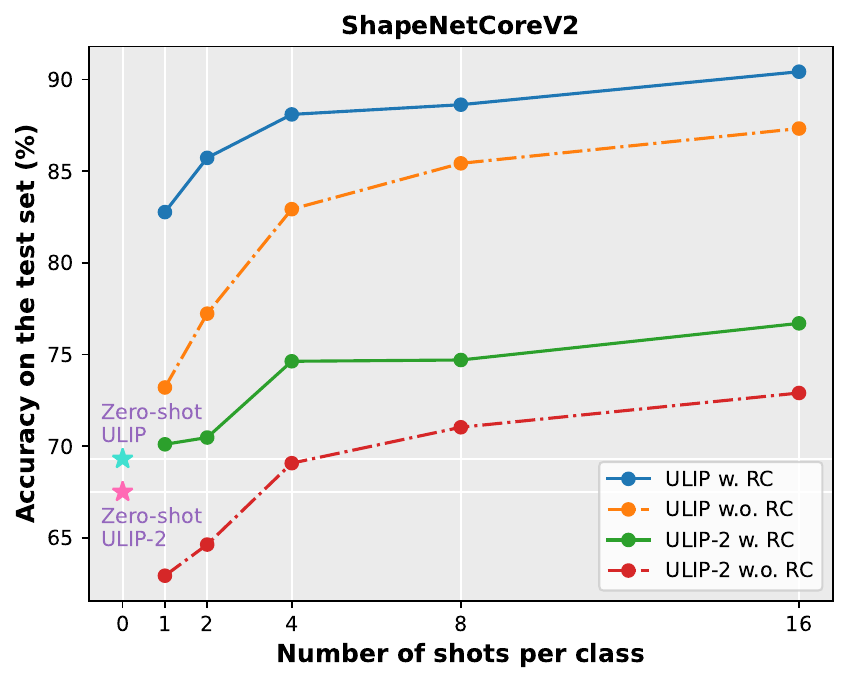}
   \caption{\textbf{Comparison of few-shot generalization}. The solid and dashed lines represent the models with and without our framework. Zero-shot performances of ULIP and ULIP-2 are marked with star symbols. The figure in the upper left presents the average results over 5 datasets.}
   \label{fig:fs_generalization}
\end{figure*}

In this setting, ULIP and ULIP-2 with (w.) and without (w.o.) our regulation constraints (RC) are compared. 
As visualized in Fig.~\ref{fig:fs_generalization}, 
the solid lines of ULIP and ULIP-2 exceed the corresponding dashed lines by clear margins average over 5 datasets, indicating 
the devised framework strengthens the 3DDG capability considerably. 
The advantages are enlarged especially for the extreme 1-shot learning, $e.g.$, +8.05\% acc. for ULIP and +5.39\% acc. for ULIP-2.
Note that in some cases, $e.g.$, on ModelNet40, ULIP-2 w.o. RC (1-shot, 66.63\%) even lags behind zero-shot ULIP-2 (71.23\%), implying that simple 
prompt tuning disturbs the well-learned representations of ULIP-2. In contrast, the developed framework brings 2.4\% 
absolute improvements over the zero-shot ULIP-2, obtaining 73.63\% acc. under the 1-shot setting.

\subsection{Ablation Study}

In this section, we examine the effectiveness of several critical components in the proposed framework 
via a series of controlled experiments. ULIP-2 is adopted as the baseline and we compare the variants on the base-to-new benchmark 
and report the harmonic mean (HM) averaged over 5 datasets. 

\textbf{Regulation constraints}. Three components in our framework are vital for generalization enhancement. 
We verify their effectiveness by adding/deleting the components on ULIP-2. 
The results in Tab.~\ref{tab:ablate_regulation_constraints_base2new}
indicate there exists a notable performance gap (6.91\%) between ULIP-2 with and without the regulation constraints. 
And the gap can be gradually narrowed down by inserting different components. For instance, ULIP-2 with the model ensemble 
constraint lifts the HM from 72.57\% to 78.89\%, a 6.32\% absolute increase. Although a single mutual agreement or 
text diversity constraint may not bring adequate gains, their combination contributes to the generalization improvement significantly, 
achieving 79.26\% that is close to the performance of the full version of our framework. 

\begin{wraptable}{r}{0.47\linewidth}
    \centering\scriptsize
    \caption{\textbf{Ablation study for the three regulation constraints in our framework}. The results are averaged on 5 datasets. 
    }
    \label{tab:ablate_regulation_constraints_base2new}
    \begin{tabular}{c c c | c c c}
       \toprule
       MAC & TDC & MEC & Base & New & HM \\
       \midrule
       \xmark & \xmark & \xmark & 77.91 & 67.91 & 72.57 \\ %
       \cmark & \xmark & \xmark & 78.02 & 68.91 & 73.18 \\ %
       \xmark & \cmark & \xmark & 78.89 & 70.63 & 74.53 \\ %
       \xmark & \xmark & \cmark & 84.54 & 72.19 & 78.89 \\ %
       \xmark & \cmark & \cmark & \textbf{84.92} & 71.17 & 77.44 \\ %
       \cmark & \xmark & \cmark & 83.19 & 73.35 & 77.96 \\ %
       \cmark & \cmark & \xmark & 83.30 & 75.59 & 79.26 \\ %
       \cmark & \cmark & \cmark & 83.18 & \textbf{76.10} & \textbf{79.48} \\
       \bottomrule
    \end{tabular}
\end{wraptable}

\textbf{Distance metrics in MAC}. The mutual agreement among the extracted features with learnable prompts 
($e.g.$ $\tilde{\textbf{h}}_T$ and $\tilde{\textbf{h}}_P$) and the general knowledge in large 3D models 
($e.g.$ $\textbf{h}_T$ and $\textbf{h}_P$)
can be implemented in different distance metrics. Here we explore their effect and report the results in Tab.~\ref{tab:ablate_mac_dist_base2new}. 
As observed, MAC with MSE distance attains the best recognition acc. on the base classes. But when incorporating 
the performances of new 3D categories into account, the same model with L1 distance demonstrates overall better generalization 
(76.10\% new acc. and 79.48\% HM). 
Thus we choose L1 as the distance metric in MAC by default. 

\begin{table*}[ht]
    \centering\scriptsize
    \caption{Ablation studies. The results are averaged over 5 datasets in the base-to-new benchmark.}
    \begin{subtable}{0.45\linewidth}
        \centering
        \caption{The distance metrics in MAC. L1: L1 norm, MSE: mean square error, Cosine: cosine distance.}
        \begin{tabular}{l | c c c }
        \toprule
        Metric & Base & New & HM \\ 
        \midrule
        Cosine & 78.63 & 73.73 & 76.10 \\ 
        MSE & \textbf{83.91} & 72.81 & 77.97 \\ 
        L1 & 83.18 & \textbf{76.10} & \textbf{79.48} \\ 
        \bottomrule
        \end{tabular}
        \label{tab:ablate_mac_dist_base2new}
    \end{subtable}
    \begin{subtable}{0.45\linewidth}
        \centering
        \caption{Here GPT-3.5 is short for GPT-3.5-turbo.}
        \begin{tabular}{l | c c c}
            \toprule
            Text Description & Base & New & HM \\ 
            \midrule
            GPT-3.5~\cite{chatgpt} & 83.30 & 71.44 & 76.92 \\ %
            GPT-4~\cite{gpt4weburl} & \textbf{83.46} & 71.55 & 77.05 \\ %
            PointLLM~\cite{xu23pointllm} & 83.27 & 73.83 & 78.27 \\ %
            Manual & 83.18 & \textbf{76.10} & \textbf{79.48} \\ %
            \bottomrule
         \end{tabular}
         \label{tab:ablate_prompt_type_base2new}
    \end{subtable}
    \label{tab:ablation_study}
\end{table*}

\textbf{Point cloud descriptions from different sources}. 
We hope to exploit flexible and diverse
text descriptions to reflect some vital characteristics
of the point clouds in different classes. 
The following experiments investigate the effect of  
the point cloud descriptions generated 
from different sources, including large language models like GPT-3.5~\cite{chatgpt}, GPT-4~\cite{gpt4weburl}, PointLLM~\cite{xu23pointllm} 
and manual templates (see Appendix for details). 
As shown in Tab.~\ref{tab:ablate_prompt_type_base2new}, 
point cloud descriptions from general-purpose LLMs, such as GPT-3.5 and GPT-4, bring decent performances 
on base classes. However, they lag behind PointLLM regarding new class recognition accuracy by a
clear margin (-2.39\% for GPT-3.5 and -2.28\% for GPT-4). We infer it is due to the fact that PointLLM has seen massive 
point cloud data and related text 
descriptions thus generates more accurate and domain-related responses. Surprisingly, 
by combining 64 simple sentences written by human beings~\cite{xue23ulip}, ULIP-2 achieves decent base class 
accuracy and the best performance on new classes, resulting in even better HM than that of ULIP-2 with LLMs' descriptions. 

\begin{figure}[ht]
    \centering
    \includegraphics[width=0.49\linewidth]{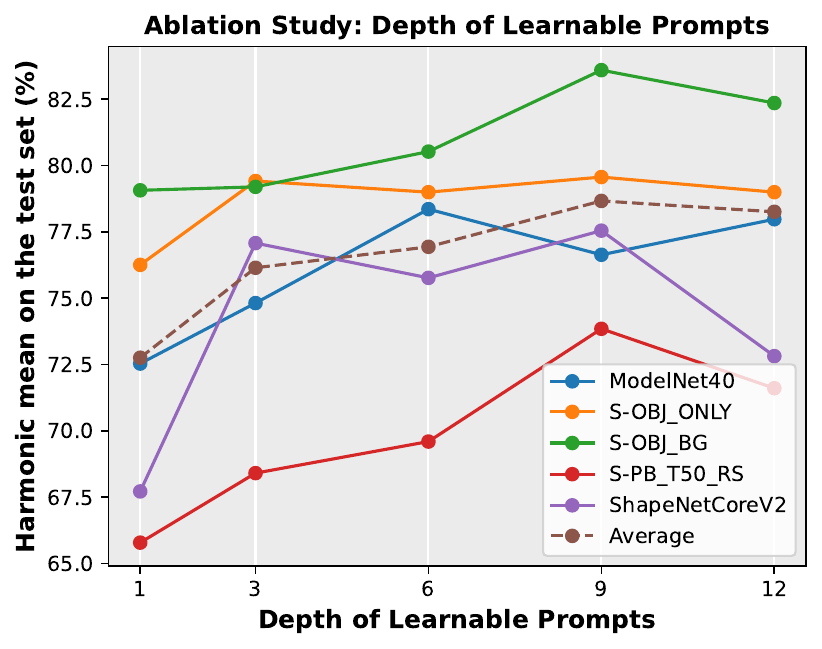}%
    \includegraphics[width=0.49\linewidth]{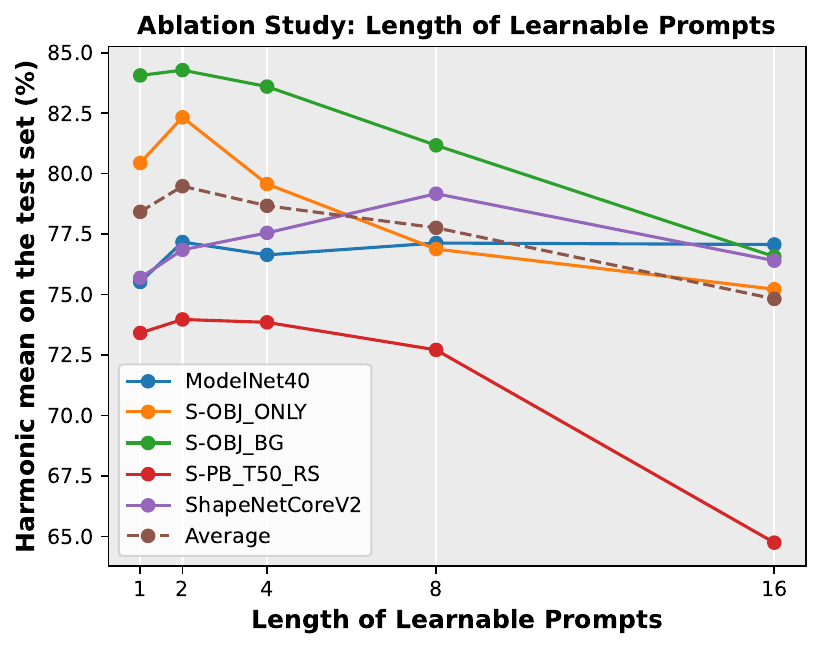}%
    \caption{\textbf{Ablation study for the prompt depth and length}. We compare the harmonic mean on five datasets of the base-to-new 
    benchmark and the average results are displayed in dashed lines.}
    \label{fig:ablate_prompt_depth_and_width_base2new}
\end{figure}

\textbf{The depth and length of learnable prompts}. Two variables that should be determined for the 
learnable prompts \{$\textit{\textbf{E}}^P$,\ $\textit{\textbf{E}}^T$\} are the depth of prompt layers $D$ and 
the length of prompt tokens $L$. 
For simplicity, the prompt depth is kept the same in the point cloud and text encoders, and similarly for the prompt length. 
We ablate the two variables and visualize the results in Fig.~\ref{fig:ablate_prompt_depth_and_width_base2new}.
In general, increasing the prompt layers promotes the harmonic mean. But it
is not always beneficial to deepen the learnable prompts, $e.g.$ ULIP-2 with $D=12$ achieves 78.26\% HM, slightly lower than 
78.67\% HM of ULIP-2 with $D=9$. 
We also find that it is not necessary to construct very long prompt tokens to achieve better generalization, $e.g.$, 
ULIP-2 with $L=2$ surpasses other variants average on 5 datasets clearly. Thus we let $D=9$ and $L=2$ by default. 

\section{Conclusion}
\label{sec:conclusion}
This paper initializes the efforts of addressing the corrupted generalization of large 
3D models when adapting to downstream 3D tasks by a comprehensive regulation framework. 
The framework enables the learnable prompts to 
actively engage with large 3D models by maximizing the mutual agreement between task-specific prediction and 
general knowledge. Consistent generalization gains are obtained over different large 3D models, suggesting the 
model-agnostic attribute of the proposed framework. We also contribute to the study of 3DDG by developing 
new and more challenging evaluation benchmarks that will drive further investigation. 
Nevertheless, this work focuses on the point cloud recognition, and we plan to discuss the segmentation and 
detection tasks in future work. 

\textbf{Limitations and Broader Impacts.} The proposed framework has demonstrated effectiveness and scalability 
on the object-level recognition task but not been validated on the scene-level tasks, such as 3D semantic segmentation 
and object detection. Different solutions may be required to handle scene-level point cloud data. On the other hand, 
when exploiting the power of LLMs to reflect critical characteristics of 3D objects, we simply ensemble multiple descriptions 
through the pooling operation, more sophisticated prompting and fusion strategy can be developed. For broader impacts, 
we are the first to investigate the generalization ability of large multi-modal 3D models, which mirrors the progress 
of the vision-language field (CLIP-based image recognition) and probably inspires a series of follow-up works. 
We do not perceive the potential negative impacts of this work. 

\begin{ack}
We thank all anonymous reviewers and area chairs for their time and valuable feedback. 
This work was partially supported by China Scholarship Council (CSC) under the Grant No. 202306360147 
and partially supported by NeurIPS 2024 Student Travel Grant. 
\end{ack}

\small{
\bibliographystyle{abbrv}
\bibliography{egbib}
}

\newpage

\appendix

\section{Appendix}

\subsection{Our New 3DDG Benchmarks}
\label{subsec:new_3ddg_benchmarks}
To our knowledge, PointDA~\cite{qin19pointdan} and Sim2Real~\cite{huang21metasets} are rare benchmarks in 3DDG.  
We perceive that existing 3DDG benchmarks may not be sufficient to cover common generalization evaluation scenarios. 
For instance, PointDA~\cite{qin19pointdan} selected 10 shared classes among three popular point cloud datasets~\cite{modelnet,dai17scannet,shapenet2015}
for generalization evaluation. Sim2Real~\cite{huang21metasets} picked 9 same categories among ShapeNet~\cite{shapenet2015} and ScanObjectNN~\cite{uy19sonn} 
and 11 shared classes in ModelNet~\cite{modelnet} and ScanObjectNN~\cite{uy19sonn}. 
Both of them emphasize the generalization among shared 3D object classes in different datasets.
But they fail to consider how to transfer to unseen 3D object classes and other out-of-distribution scenarios. 
To alleviate the drawbacks, we develop three new benchmarks, including
base-to-new, cross-dataset, and few-shot generalization to enrich 3DDG evaluation and drive future research. 

\subsubsection{Construction}
\textbf{Base-to-new Class Benchmark.} Inspired by the evaluation settings in 2D vision~\cite{zhou22cocoop}, 
we take five 3D datasets, including ModelNet40~\cite{modelnet}, three variants of 
ScanObjectNN~\cite{uy19sonn} (S-PB\_T50\_RS, S-OBJ\_BG, S-OBJ\_ONLY), and ShapeNetCoreV2~\cite{shapenet2015} to construct the 
base-to-new benchmark. Each of them are equally divided into two halves object classes, called base and new classes. 
Specifically, the first half is regarded as base and the second half is treated as new. 
The train, val, test sets are split for the base classes while the new classes only serve the test purpose. 
The splitting adopts the official standards of the released datasets if available (e.g., ModelNet40 and ShapeNetCoreV2), 
otherwise (e.g., 3 variants of ScanObjectNN) we randomly selected 20\% samples 
in the original training set to be the validation set and keep the remaining 80\% as new training set. 
The original test set is unchanged to serve as the test set. 
This arrangement allows the model to be trained on base classes then tested on new classes so that can measure 
its generalization on unseen new categories and data. 

\begin{table}[ht]
    \centering\scriptsize
    \caption{Statistics of the Base-to-New benchmark.}
    \begin{tabular}{l r r c r r c r r c r r c r r}
        \toprule
        \multirow{2}{*}{Item} & \multicolumn{2}{c}{ModelNet40} & & \multicolumn{2}{c}{S-OBJ\_ONLY} & & \multicolumn{2}{c}{S-OBJ\_BG} & &
        \multicolumn{2}{c}{S-PB\_T50\_RS} & & \multicolumn{2}{c}{ShapeNetCoreV2} \\
        \cline{2-3}\cline{5-6}\cline{8-9}\cline{11-12}\cline{14-15}
        & base & new & & base & new & & base & new & & base & new & & base & new\\
        \midrule
        \#classes & 20 & 20 & & 8 & 7 & & 8 & 7 & & 8 & 7 & & 28 & 27 \\
        \#train & 4,084 & N/A & & 1,055 & N/A & & 1,062 & N/A & & 5,321 & N/A & & 23,584 & N/A \\
        \#val & 1,028 & N/A & & 281 & N/A & & 274 & N/A & & 1,268 & N/A & & 3,401 & N/A \\
        \#test & 1,202 & 1,266 & & 352 & 229 & & 352 & 229 & & 1,741 & 1,141 & & 6,960 & 3,301 \\
        \bottomrule
    \end{tabular}
    \label{tab:stat_base2new}
\end{table}

\textbf{Cross-Dataset Benchmark.} 
Previous Sim2Real~\cite{huang21metasets} and PointDA~\cite{qin19pointdan} evaluate the generalization across 
totally shared categories in different datasets. But in real world the source and target domains do not necessarily
have common classes. In this work, we construct the cross-dataset benchmark to expand the generalization evaluation to 
broader scopes, which incorporates two newly introduced settings, \emph{OOD generalization} and \emph{data corruption}, and 
another \emph{Sim2Real}~\cite{huang21metasets} and \emph{PointDA}~\cite{qin19pointdan}. 
Note that the last two datasets are released by other researchers for 3DDG evaluation and we just follow their default settings.  

\emph{OOD Generalization.} 
In this setting, the source and target domain may not necessarily share common categories. 
Also, the number of object classes can be different.
ShapeNetCoreV2~\cite{shapenet2015} is arranged as the source dataset while ModelNet40~\cite{modelnet}, the three variants of ScanObjectNN~\cite{uy19sonn} and Omni3D~\cite{OmniObject3D} serve as the target evaluation places. 
Apparently, this design is specially for large 3D models that have open-vocabulary recognition ability since 
traditional 3DDG methods like MetaSets~\cite{huang21metasets}, PDG~\cite{wei22pdg} are only able to recognize a fixed set of 
point cloud classes.

\emph{Data Corruption.} %
Point cloud corruption is inevitable due to irregular geometry structures, inaccurate sensors or processing errors. 
Existing 3DDG benchmarks fail to take this factor into account.  
We utilize the off-the-shelf point cloud corruption dataset ModelNet-C~\cite{ren22modelnet-c} as 
a place to measure the robustness and generalization of point cloud recognition methods against common data corruptions, 
such as losing local parts, global noise, $etc$. 

\begin{table*}[ht]
    \centering\scriptsize
    \caption{Statistics of the Cross-Dataset benchmark.}
    \begin{subtable}{\linewidth}\
        \centering
        \caption{\emph{OOD Generalization}}
        \begin{tabular}{l c r r r r r r}
            \toprule
            \multirow{2}{*}{Item} & Source & & \multicolumn{5}{c}{Target} \\\cline{2-2}\cline{4-8}
                                    & ShapeNetCoreV2 & & ModelNet40 & S-OBJ\_ONLY & S-OBJ\_BG & S-PB\_T50\_RS & Omni3D \\
            \midrule
            \#classes & 55 &  & 40 & 15  & 15 & 15 & 216 \\
            \#train & 35,708 &  & 7,861 & 1,847  & 1,847 & 9,132 & N/A \\
            \#val & 5,158 &  & 1,979 & 462  & 462 & 2,284 & N/A \\
            \#test & 10,261 &  & 2,468 & 581  & 581 & 2,882 & 5,910 \\
            \bottomrule
        \end{tabular}
        \label{tab:stat_xset_dg}
    \end{subtable}
    \begin{subtable}{\linewidth}
        \vspace{9pt}
        \centering
        \caption{\emph{Data Corruption}}
        \begin{tabular}{l r r r r r r r}
            \toprule
            Item & Add Global & Add Local & Drop Global & Drop Local & Rotate & Scale & jitter \\
            \midrule
            \#classes & 40 & 40 & 40 & 40 & 40 & 40 & 40 \\
            \#test & 2,468 & 2,468 & 2,468 & 2,468 & 2,468 & 2,468 & 2,468 \\
            \bottomrule
        \end{tabular}
    \end{subtable}
    \begin{subtable}{\linewidth}
        \vspace{9pt}
        \centering
        \caption{\emph{Sim-to-Real}}
        \begin{tabular}{l r c r r r }
            \toprule
            \multirow{2}{*}{Item} & Source & & \multicolumn{3}{c}{Target} \\\cline{2-2}\cline{4-6}
                                & ModelNet &  & S-OBJ\_ONLY & S-OBJ\_BG & S-PB\_T50\_RS \\
            \midrule
            \#classes & 11 &  & 11  & 11 & 11  \\
            \#train & 4,844 &  & N/A  & N/A & 9,132  \\
            \#test & 972 &  & 475  & 475 & 2,882  \\
            \bottomrule
        \end{tabular}
        \label{tab:stat_xset_sim2real}
    \end{subtable}
    \quad
    \begin{subtable}{\linewidth}
        \vspace{9pt}
        \centering
        \caption{\emph{PointDA}}
        \begin{tabular}{l r r r}
            \toprule
            Item & ModelNet & ShapeNet & ScanNet \\
            \midrule
            \#classes & 10 & 10 & 10 \\
            \#train & 4,183 & 17,378 & 6,110 \\
            \#test & 856 & 2,492 & 1,769 \\
            \bottomrule
        \end{tabular}
        \label{tab:stat_xset_pointda}
    \end{subtable}
    \label{tab:stat_xset}
\end{table*}

\textbf{Few-shot Benchmark.} This benchmark incorporates same datasets with those in the 
\emph{Base-to-new Class} Benchmark. But we do not distinguish the base and new classes 
and treat them as a whole. During prompt learning on each dataset of this benchmark, 
we randomly sample 1, 2, 4, 8 and 16 shots from each category to tune the learnable 
prompts, then observe the generalization on the whole test set. 

\begin{table}[ht]
    \centering\footnotesize
    \caption{Statistics of the Few-shot benchmark.}
    \begin{tabular}{l r r r r r }
        \toprule
        Item & ModelNet40 & S-OBJ\_ONLY & S-OBJ\_BG & S-PB\_T50\_RS & ShapeNetCoreV2 \\
        \midrule
        \#classes & 40 & 15  & 15 & 15 & 55 \\
        \#train & 7,861 & 1,847  & 1,847 & 9,132 & 35,708 \\
        \#val & 1,979 & 462  & 462 & 2,284 & 5,158 \\
        \#test & 2,468 & 581  & 581 & 2,882 & 10,261 \\
        \bottomrule
    \end{tabular}
    \label{tab:stat_fewshot}
\end{table}

\subsubsection{The Evaluation Settings}
\label{subsec:general_eval_settings}

\textbf{Base-to-New}. This benchmark includes five point cloud datasets as described above. 
Each dataset is equally split into base (known) and new (unseen) classes. We compute the recognition accuracy (Acc.)
on the two types of classes, respectively. Although the accuracy on the new classes reflects how well a model can learn from 
known point cloud categories to generalize other unseen data, we also want decent accuracy on the base classes. 
Consequently, the harmonic mean (HM) between the accuracies of base and new classes is chosen to balance the two factors. 
Similar evaluation settings are also adopted in vision-language community~\cite{zhou22cocoop,lu22prompt,chen23plot,khattak23maple,khattak23promptsrc,zhu23prograd,li24promptkd}.
Note that the models trained on a fixed set of point cloud categories cannot be evaluated on this benchmark since they 
do not have the open-vocabulary recognition ability, such as IDPT~\cite{zha23idpt}, Point-PEFT~\cite{tang24pointpeft}, 
DAPT~\cite{zhou24dapt}. 

\textbf{Cross-Dataset}. This benchmark has four types of evaluation settings and we will focus on the first two settings which 
are, \emph{OOD generalization} and \emph{data corruption}, newly introduced in 3DDG by us. 
The last two correspond to the default settings in PointDA~\cite{qin19pointdan} and Sim-to-Real~\cite{huang21metasets}. 
For \emph{OOD generalization}, models are trained on the source domain and directly transferred to the target domains for evaluation.  
Recognition accuracy is a major metric. We will average the accuracies across five target domains to measure the 
final generalization ability. 
For \emph{data corruption}, models are trained on clean ModelNet~\cite{modelnet} and tested on various point cloud corruption scenarios 
in ModelNet-C\cite{ren22modelnet-c}, such as add global noises, losing local parts, geometry transformations, $etc$. 
The average accuracies on 7 types of atomic corruptions are computed to measure the model robustness against point cloud data
corruption. 

\textbf{Few-Shot}. This setting inspects the model generalization in extremely low-data regime, where only a few samples of each class 
are offered to train a model then it is evaluated on the whole test set. Here we take 1, 2, 4, 8, and 16 shots for the model training and 
the recognition accuracy on the whole test set is compared. 

We insert the proposed explicit constraints to large 3D models, then conduct lightweight prompt tuning  
on downstream 3D tasks, finally observe whether positive gains will appear in above 3DDG evaluation settings
compared to the same prompt learning but without our regulation framework. 

\subsection{Implementation Details}
In gaussian model weighting, we take $\mu = 15$ and $\sigma = 1$. 
To avoid store $e$ separate copies of the model parameters, 
we implement Eq.~\ref{eq:model_ensemble_constraint} by iteratively adding 
current and previous epoch of weighted model parameters. Note the learnable parameters are randomly initialized before training. 
The design ensures our model absorbing prior knowledge and reduces disk consumption effectively. 
Both ULIP and ULIP-2 exploit the Point-BERT~\cite{yu22pointbert} as the backbone. 

We use two RTX 4090 GPUs to run the experiments. 
On the base-to-new benchmark, we conduct prompt learning for 20 epochs. On the cross-dataset and few-shot benchmark, 
models are trained for 50 epochs, and the prompt depth $D$ and length $L$ are set to 12 and 4, respectively.
For the evaluation on ModelNet-C, we report the results when the corruption severity is 2.
The 64 manual templates and other 3D object classes descriptions generated by LLMs are available at our provided codebase. 

\subsection{Additional Results and Analysis}

\subsubsection{Base-to-new Class Generalization}
Table~\ref{tab:base2new_generalization_with_variance} enriches the 
\emph{base-to-new class} generalization evaluation by reporting the mean 
and standard deviation of three runnings on the \emph{base-to-new class} benchmark. 
The standard deviation in the bracket follows the mean. 
Note that ULIP and ULIP-2 with our regulation framework have much smaller deviation on novel classes, $e.g.$, 2.38\% vs 3.77\%, 
suggesting better generalization and robustness 
against the variations between seen and unseen point cloud data. 

\begin{table*}[ht]\scriptsize
    \centering
    \caption{\textbf{Base-to-new class generalization comparison for representative large 3D models based on prompt learning}. 
    Each number here is the mean of three runnings. 
    Base: base class accuracy (in \%, same below). New: new class accuracy. HM: harmonic mean of base and new class accuracy. 
    +\textbf{RC} demonstrates the models with our regulation constraint framework. 
    }
    \begin{subtable}{0.47\linewidth}
       \centering
       \caption{\textbf{Average over 5 datasets}}
       \begin{tabular}{l c c | c}
       \toprule
       Method & Base & New & HM \\
       \midrule
       P-CLIP~\cite{zhang22pointclip} & 75.66 & 23.45 & 35.80 \\
       P-CLIP2~\cite{zhu23pointclip2} & 74.11 & 37.84 & 50.10 \\ %
       \midrule
       ULIP~\cite{xue23ulip} & 77.32(1.41) & 49.01(3.77) & 59.99 \\
       +\textbf{RC}(Ours) & \textbf{82.19}(1.22) & \textbf{61.93}(2.38) & \textbf{70.64} \\
       \midrule
       ULIP-2~\cite{xue23ulip2} & 77.91(0.95) & 67.91(3.75) & 72.57 \\ %
       +\textbf{RC}(Ours) & \textbf{83.18}(0.62) & \textbf{76.10}(1.14) & \textbf{79.48} \\
       \bottomrule
       \end{tabular}
       \label{tab:base2new_avg_five_datasets_with_variance}
    \end{subtable}
    \quad
    \begin{subtable}{0.47\linewidth}
        \centering
       \caption{ModelNet40}
       \begin{tabular}{l c c | c}
       \toprule
       Method & Base & New & HM \\
       \midrule
       P-CLIP~\cite{zhang22pointclip} & 93.23 & 20.22 & 33.23\\
       P-CLIP2~\cite{zhu23pointclip2} & 93.98 & 45.21 & 61.05\\
       \midrule
       ULIP~\cite{xue23ulip} & 92.80(0.93) & 50.07(3.52) & 65.05 \\
       +\textbf{RC}(Ours) & \textbf{95.03}(0.52) & \textbf{55.27}(3.03) & \textbf{69.89} \\
       \midrule
       ULIP-2~\cite{xue23ulip2} & 91.77(0.41) & 56.47(2.78) & 69.92 \\ %
       +\textbf{RC}(Ours) & \textbf{95.30}(0.36) & \textbf{64.83}(0.26) & \textbf{77.17}\\
       \bottomrule
       \end{tabular}
       \label{tab:base2new_mn40_with_variance}
    \end{subtable}
    \begin{subtable}{0.47\linewidth}
        \vspace{9pt}
        \centering
       \caption{S-PB\_T50\_RS}
       \begin{tabular}{l c c | c}
       \toprule
       Method & Base & New & HM \\
       \midrule
       P-CLIP~\cite{zhang22pointclip} & 61.25 & 19.87 & 30.01\\
       P-CLIP2~\cite{zhu23pointclip2} & 56.84 & 29.92 & 39.20\\
       \midrule
       ULIP~\cite{xue23ulip} & 56.73(0.84) & 25.80(2.33) & 35.47 \\
       +\textbf{RC}(Ours) & \textbf{64.20}(0.99) & \textbf{49.17}(2.55) & \textbf{55.69} \\
       \midrule
       ULIP-2~\cite{xue23ulip2} & 66.40(1.39) & 66.47(2.40) & 66.43 \\
       +\textbf{RC}(Ours) & \textbf{73.67}(0.56) & \textbf{74.27}(1.27) & \textbf{73.97}\\
       \bottomrule
       \end{tabular}
       \label{tab:base2new_so_pb_t50_rs_with_variance}
    \end{subtable}
    \quad
    \begin{subtable}{0.47\linewidth}
        \vspace{9pt}
        \centering
       \caption{S-OBJ\_BG}
       \begin{tabular}{l c c | c}
       \toprule
       Method & Base & New & HM \\
       \midrule
       P-CLIP~\cite{zhang22pointclip} & 72.82 & 23.00 & 34.96 \\
       P-CLIP2~\cite{zhu23pointclip2} & 70.07 & 35.08 & 46.75\\
       \midrule
       ULIP~\cite{xue23ulip} & 73.20(2.32) & 47.17(1.76) & 57.37 \\
       +\textbf{RC}(Ours) & \textbf{79.47}(1.92) & \textbf{55.20}(2.89) & \textbf{65.15} \\
       \midrule
       ULIP-2~\cite{xue23ulip2} & 77.00(1.04) & 83.27(3.76) & 80.01 \\
       +\textbf{RC}(Ours) & \textbf{80.10}(0.99) & \textbf{88.93}(0.24) & \textbf{84.28}\\
       \bottomrule
       \end{tabular}
       \label{tab:base2new_so_obj_bg_with_variance}
    \end{subtable}
    \begin{subtable}{0.47\linewidth}
       \vspace{9pt}
       \centering
       \caption{S-OBJ\_ONLY}
       \begin{tabular}{l c c | c}
       \toprule
       Method & Base & New & HM \\
       \midrule
       P-CLIP~\cite{zhang22pointclip} & 76.23 & 20.23 & 31.97\\
       P-CLIP2~\cite{zhu23pointclip2} & 71.40 & 44.39 & 54.74\\
       \midrule
       ULIP~\cite{xue23ulip} & 74.13(0.60) & 50.80(7.71) & 60.29 \\
       +\textbf{RC}(Ours) & \textbf{79.23}(1.44) & \textbf{65.93}(2.21) & \textbf{71.97} \\
       \midrule
       ULIP-2~\cite{xue23ulip2} & 78.60(0.37) & 76.27(4.72) & 77.42 \\
       +\textbf{RC}(Ours) & \textbf{83.60}(0.37) & \textbf{81.10}(2.69) & \textbf{82.33}\\
       \bottomrule
       \end{tabular}
       \label{tab:base2new_so_obj_only_with_variance}
    \end{subtable}
    \quad
    \begin{subtable}{0.47\linewidth}
       \vspace{9pt}
       \centering
       \caption{ShapeNetCoreV2}
       \begin{tabular}{l c c | c}
       \toprule
       Method & Base & New & HM \\
       \midrule
       P-CLIP~\cite{zhang22pointclip} & 74.78 & 33.92 & 46.61\\
       P-CLIP2~\cite{zhu23pointclip2} & 78.27 & 34.58 & 47.97\\
       \midrule
       ULIP~\cite{xue23ulip} & 89.73(2.38) & 71.20(3.51) & 79.40 \\
       +\textbf{RC}(Ours) & \textbf{93.03}(1.23) & \textbf{84.10}(1.24) & \textbf{88.34} \\
       \midrule
       ULIP-2~\cite{xue23ulip2} & 75.80(1.56) & 57.07(5.09) & 65.11 \\ %
       +\textbf{RC}(Ours) & \textbf{83.23}(0.80) & \textbf{71.37}(1.23) & \textbf{76.85}\\
       \bottomrule
       \end{tabular}
       \label{tab:base2new_snv2_with_variance}
    \end{subtable}
    \label{tab:base2new_generalization_with_variance}
 \end{table*}

\subsubsection{Cross-dataset Generalization}
\emph{Sim-to-Real} evaluation measures the 3D domain generalization from simulating data to real world. 
This evaluation was first introduced by MetaSets~\cite{huang21metasets} then followed by PDG~\cite{wei22pdg}. 
In this setting, ModelNet~\cite{modelnet} and ShapeNet~\cite{shapenet2015} are regarded as synthetic point cloud and ScanObjectNN is constructed based on 
real-scan data. 
Here we implement the proposed regulation framework upon P-CLIP2~\cite{zhu23pointclip2}, ULIP~\cite{xue23ulip}, ULIP-2~\cite{xue23ulip2} 
and compare with prior state-of-the-art. Note that MetaSets and PDG use the whole training set in the source domain for 
supervised learning, whereas our methods only exploit 16-shot prompt tuning. As Tab.~\ref{tab:xset_sim2real_generalization} shows, our framework brings consistent generalization improvement on different large 3D models, 
+0.52\% for P-CLIP2, +5.88\% for ULIP and +2.48\% for ULIP-2 average on 6 datasets. 
The enhanced ULIP-2 by the devised framework also outreaches prior best-performing PDG-D by 5.82\%, demonstrating better generalization 
to real-world point cloud data. 

\begin{table*}[ht]
   \centering\scriptsize
   \caption{\textbf{Comparison of cross dataset generalization on \textit{Sim-to-Real}}. There are two evaluation settings, 
   MN\_11 $\rightarrow$ SONN\_11, SN\_9 $\rightarrow$ SONN\_9. 
   The left side of $\rightarrow$ stands for simulating data and the right side indicates real-world data.  
   11 classes are shared between ModelNet and ScanObjectNN, 9 classes are common in ShapeNet and ScanObjectNN. 
   In the experiments, a point cloud contains 2048 points. 
   -P: PointNet, -D: DGCNN.}
   \begin{tabular}{l c c c c c c c c}
      \toprule
      \multirow{2}{*}{Method} & \multicolumn{3}{c}{MN\_11 $\rightarrow$ SONN\_11} & & \multicolumn{3}{c}{SN\_9 $\rightarrow$ SONN\_9} & \multirow{2}{*}{\textbf{Avg.}} \\\cline{2-4}\cline{6-8}
             & OBJ & OBJ\_BG & PB\_T50\_RS & & OBJ & OBJ\_BG & PB\_T50\_RS\\
      \midrule
      MetaSets-P~\cite{zhu23pointclip2} & 60.3 & 52.4 & 47.4 & & 51.8 & 44.3 & 45.6 & 50.3 \\ %
      MetaSets-D~\cite{zhu23pointclip2} & 58.4 & 59.3 & 48.3 & & 49.8 & 47.4 & 42.7 & 51.0\\ %
      PDG-P~\cite{wei22pdg} & 67.6 & 58.5 & 56.6 & & 57.3 & 51.3 & 51.3 & 57.1 \\ %
      PDG-D~\cite{wei22pdg} & 65.3 & 65.4 & 55.2 & & 59.1 & 59.3 & 51.0 & 59.2 \\ %
      \midrule
      P-CLIP2~\cite{zhu23pointclip2} & 18.67(1.68) & \textbf{15.57}(3.23) & \textbf{15.63}(1.63) &  & 53.00(3.06) & 47.83(1.84) & 35.83(0.19) & 31.09(1.94) \\
      +\textbf{RC}(Ours) & \textbf{19.23}(4.00) & 15.50(3.88) & 14.37(2.57) & & \textbf{56.60}(3.70) & \textbf{47.83}(3.80) & \textbf{36.10}(2.81) & \textbf{31.61}(3.46) \\
      \midrule
      ULIP~\cite{xue23ulip} & 21.60(2.50) & 18.03(2.03) & 13.63(1.51) & & 54.83(1.66) & 54.17(2.46) & 40.87(1.27) & 33.86(1.91) \\
      +\textbf{RC}(Ours) & \textbf{29.90}(1.15) & \textbf{24.07}(2.03) & \textbf{18.87}(2.52) & & \textbf{63.13}(0.74) & \textbf{58.87}(0.49) & \textbf{43.60}(1.51) & \textbf{39.74}(1.41) \\
      \midrule
      ULIP-2~\cite{xue23ulip2} & 62.73(0.95) & 68.23(0.86) & 52.83(1.10) & & 66.90(2.77) & 70.50(2.48) & 54.03(2.75) & 62.54(1.82) \\ %
      +\textbf{RC}(Ours) & \textbf{68.43}(1.07) & \textbf{69.47}(0.95) & \textbf{55.30}(2.00) & & \textbf{65.83}(1.35) & \textbf{72.53}(0.47) & \textbf{58.57}(1.17) & \textbf{65.02}(1.17) \\ %
      \bottomrule
   \end{tabular}
   \label{tab:xset_sim2real_generalization}
\end{table*}

\emph{PointDA} is a 3D domain adaptation benchmark introduced by PointDAN~\cite{qin19pointdan}, which includes 
6 evaluation settings as displayed in Tab.~\ref{tab:xset_pointda_generalization}. 
Previous methods like MetaSets~\cite{huang21metasets}, PDG~\cite{wei22pdg}, I-OODG~\cite{zhang24invariantoodg} exploit the full training data in each setting while we adopt few-shot learning (16 shots). 
The results suggest the proposed framework contributes to the enhanced domain adaptation significantly, $e.g.$, 
almost 10\% absolute improvements for ULIP-2 that leads prior state-of-the-art I-OODG by 6.94\%.

\begin{table*}[ht]
   \centering\scriptsize
   \caption{\textbf{Comparison of cross-dataset generalization on \textit{PointDA}}. M: ModelNet, S: ShapeNet, S*: ScanNet. The last column is the average over 6 evaluation settings.}
   \label{tab:xset_pointda_generalization}
   \begin{tabular}{l c c c c c c c}
      \toprule
      Method & M $\rightarrow$ S & M $\rightarrow$ S* & S $\rightarrow$ M & S $\rightarrow$ S* & S* $\rightarrow$ M 
      & S* $\rightarrow$ S & \textbf{Avg.} \\
      \midrule
      P-DAN~\cite{qin19pointdan} & 64.2 & 33.0 & 47.6 & 33.9 & 49.1 & 64.1 & 48.7 \\
      MetaSets~\cite{huang21metasets} & 86.0 & 52.3 & 67.3 & 42.1 & 69.8 & 69.5 & 64.5 \\
      PDG~\cite{wei22pdg} & 85.6 & 57.9 & 73.1 & 50.0 & 70.3 & 66.3 & 67.2 \\
      I-OODG~\cite{zhang24invariantoodg} & 83.7 & 56.4 & 71.7 & 57.6 & 69.5 & 73.5 & 67.8\\ %
      \midrule
      P-CLIP2~\cite{zhu23pointclip2} & 40.53 & 26.40 & 31.33 & 35.57 & 16.30 & 24.97 & 29.18 \\
      +\textbf{RC}(Ours) & \textbf{43.27} & \textbf{37.20} & \textbf{32.47} & \textbf{36.97} & \textbf{21.70} & \textbf{43.90} & \textbf{35.92} \\
      \midrule
      ULIP~\cite{xue23ulip} & 74.33(8.63) & 38.23(2.12) & 35.17(3.99) & 36.17(5.67) & 24.70(5.16) & 60.67(4.72) & 44.88(5.05) \\
      +\textbf{RC}(Ours) & \textbf{78.80}(1.49) & \textbf{41.63}(0.87) & \textbf{43.03}(4.28) & \textbf{41.60}(4.02) & \textbf{25.23}(4.68) & \textbf{63.60}(9.01) & \textbf{48.98}(4.06) \\
      \midrule
      ULIP-2~\cite{xue23ulip2} & 84.80(2.69) & 48.10(2.13) & 83.20(4.17) & 42.00(4.18) & 60.43(4.83) & 70.50(6.22) & 64.84(4.04) \\ %
      +\textbf{RC}(Ours) & \textbf{89.00}(1.18) & \textbf{51.37}(1.03) & \textbf{89.87}(2.38) & \textbf{49.57}(2.50) & \textbf{85.57}(3.80) & \textbf{83.07}(4.21) & \textbf{74.74}(2.52) \\ %
      \bottomrule
   \end{tabular}
\end{table*}

\subsubsection{The Role of MEC}
Selecting the best checkpoint through a validation set is a common way. In theory, this greedy strategy favors highest 
performances on the downstream tasks, which means the small number of learnable prompts/parameters are well adapted to these 
tasks. It is equivalent to our framework without \emph{Model Ensemble Constraint} (MEC).

However, purely optimizing the small number of learnable prompts toward target tasks will inevitably hinder the generalization 
ability of the large 3D models, as we analyzed in the paper.

We also provide the ablation study to this problem, as the results in Tab.~\ref{tab:ablate_mec} indicates, the method without MEC has slightly lower 
accuracy on new classes (75.59\% vs 76.10\%) and harmonic mean. However, when removing the factors of MAC and TDC, the role of 
MEC becomes prominent. It raise the overall performance remarkably, especially for unseen new classes (5.28\% absolute points).

\begin{table*}[ht]
    \centering\scriptsize
    \caption{\textbf{Ablation study for the framework without model ensembling constraint}. The results are averaged on 5 datasets. MAC: mutual agreement constraint, TDC: text diversity constraint, MEC: model ensemble constraint. HM: harmonic mean of the Base and New class accuracies.}
    \label{tab:ablate_mec}
    \begin{tabular}{l c c c c c c c}
        \toprule
        \multirow{2}{*}{Model} & \multicolumn{3}{c}{Variant} & & \multicolumn{3}{c}{Performance} \\\cline{2-4}\cline{6-8}
         & MAC & TDC & MEC & & Base & New & HM \\
        \midrule
        \multirow{4}{*}{ULIP-2} & \xmark & \xmark & \xmark & & 77.91 & 67.91 & 72.51 \\
        & \xmark & \xmark & \cmark & & 82.42 & 73.19 & 77.53 \\\cline{2-8}
        & \cmark & \cmark & \xmark & & 83.30 & 75.59 & 79.26 \\
        & \cmark & \cmark & \xmark & & 83.18 & 76.10 & 79.48 \\
        \bottomrule
    \end{tabular}
\end{table*}

\subsubsection{Running Time}
We compare the training time between our method and the baselines. The results are shown in Tab.~\ref{tab:training_time}. The proposed method consumes a similar amount of time per epoch compared to the baseline, with a slight increase due to 
the inclusion of our framework.

\begin{table*}[ht]
    \centering\scriptsize
    \caption{\textbf{Running time comparison of a strong baseline ULIP-2 and the proposed approach}. We conduct prompt learning 
    based on ULIP-2 for 20 epochs on the base-to-new class benchmark, and the experiments are run three times with different 
    seeds. The settings are consistent with those in the main paper. Time is counted in \emph{seconds} for all 20 epochs using a RTX 4090.}
    \label{tab:training_time}
    \begin{tabular}{l c c c c c c c}
        \toprule
        \multirow{2}{*}{Method}  & \multirow{2}{*}{seed} & \multicolumn{5}{c}{Dataset} & \multirow{2}{*}{\textbf{Avg.}} \\\cline{3-7}
         &  & MN40 & S-PB\_T50\_RS & S-OBJ\_BG &S S-OBJ\_ONLY & SNV2 & \\
         \midrule
         \multirow{3}{*}{ULIP-2} & 1 & 132 & 106 & 48 & 53 & 307 & 129.2 \\
         & 2 & 132 & 106 & 48 & 53 & 305 & 128.8 \\
         & 3 & 133 & 108 & 48 & 51 & 305 & 129.6 \\
        \midrule
        \multirow{3}{*}{~\textbf{+RC}(Ours)} & 1 & 159 & 112 & 60 & 60 & 344 & 147.0 \\
         & 2 & 159 & 114 & 60 & 59 & 345 & 147.4 \\
         & 3 & 159 & 113 & 59 & 60 & 345 & 147.2 \\
        \bottomrule
    \end{tabular}
\end{table*}

The number of learnable parameters of our framework is 16,896 while full fine-tuning ULIP-2 has 82.3M learnable parameters (only in text and 3D encoder). According to the reported details of ULIP-2, pre-training on Objaverse~\cite{Objaverse} utilizes 8 A100 GPUs and takes 1.5 days. So full fine-tuning ULIP-2 is also expensive.

\subsubsection{Scalability}
We further test our framework on a larger dataset named Objaverse-LVIS and the results are promising. This dataset is a subset of the recently released Objaverse and only serves as a test set (target domain). Objaverse-LVIS contains 46,205 point clouds distributed in 1,156 classes, and some classes only have a single object, posing great challenges to existing point cloud recognition methods. In the experiments, we select representative ULIP and ULIP-2 as baselines and compare them with the models with our regulation framework.
The results in the Tab.~\ref{tab:scalability} verify the proposed approach can also bring considerable gains (+3.27\% absolute points for ULIP-2) on such a larger and challenging dataset.

\begin{table*}[ht]
    \centering\scriptsize
    \caption{\textbf{Analysis of scalability}. }
    \label{tab:scalability}
    \begin{tabular}{l c c c}
        \toprule
        \multirow{2}{*}{Method} & \textbf{Source} & & \textbf{Target} \\\cline{2-2}\cline{4-4}
        & ShapeNetV2 & & Objaverse-LVIS \\
        \midrule
        ULIP & 87.33(0.95) & & 0.83(0.05) \\
        ~\textbf{+RC}(Ours) & \textbf{90.43}(0.86) & & \textbf{1.10}(0.08) \\
        \midrule
        ULIP-2 & 76.70(1.37) & & 14.80(0.22) \\
        ~\textbf{+RC}(Ours) & \textbf{76.70}(1.59) & & \textbf{18.07}(0.49) \\
        \bottomrule
    \end{tabular}
\end{table*}

\end{document}